\documentclass[10pt,twocolumn,letterpaper]{article}

\usepackage[dvipsnames,table]{xcolor}
\usepackage{iccv}
    \makeatletter
\@namedef{ver@everyshi.sty}{}
\makeatother
\usepackage{times}
\usepackage{epsfig}
\usepackage{graphicx}
\usepackage{amsmath}
\usepackage{amssymb}
\usepackage[accsupp]{axessibility}  %
\usepackage[inline]{enumitem}
\usepackage{booktabs}
\usepackage{tabularx}
\usepackage{caption}
\usepackage{subcaption}
\usepackage{multirow}
\usepackage{svg}
\usepackage{pagecolor}%
\usepackage{cuted}
\usepackage{tabu}
\usepackage{pifont}  %
\usepackage{times}
\usepackage{pgffor}
\usepackage[export]{adjustbox}

\definecolor{lightGray}{RGB}{230,230,230}
\definecolor{mediumGray}{RGB}{200,200,200}
\definecolor{darkGray}{RGB}{160,160,160}

\DeclareMathOperator*{\argmin}{arg\,min}

\newcommand{\cmark}{\ding{51}}%
\newcommand{\xmark}{\ding{55}}%

\newcommand{\comment}[1]{}

\newcommand{\TODO}[1]{{\color{red}{\bf [TODO: #1]}}}
\newcommand{\et}[1]{{\color{Magenta}{#1}}}
\newcommand{\ET}[1]{{\color{Magenta}{\bf [E: #1]}}}

\newcommand{\PF}[1]{{\color{MidnightBlue}{\bf [P: #1]}}}
\newcommand{\mt}[1]{{\color{OliveGreen}{#1}}}
\newcommand{\MT}[1]{{\color{OliveGreen}{\bf [M: #1]}}}

\renewcommand{\TODO}[1]{}
\renewcommand{\et}[1]{#1}
\renewcommand{\ET}[1]{}

\renewcommand{\PF}[1]{}
\renewcommand{\mt}[1]{#1}
\renewcommand{\MT}[1]{}

\newcommand{\ours}{GECCO}
\newcommand{\sigmamax}{\sigma_{\textrm{max}}}
\newcommand{\pointflow}{PointFlow}
\newcommand{\diffpointflow}{DPM} %
\newcommand{\cflow}{C-flow}
\newcommand{\shapegf}{ShapeGF} %
\newcommand{\pvd}{PVD} %
\newcommand{\lion}{LION} %

\newcommand{\width}{1cm}
\newcommand{\height}{1cm}

\newcommand{\mytilde}{{\raise.17ex\hbox{$\scriptstyle\sim$}}}

\newcommand{\set}[1]{\{ #1 \}} %

\newcommand{\dimnn}{d_\mathrm{nn}}

\newcommand{\Section}[1]{Sec.~\ref{sec:#1}}

\newcommand{\first}[1]{{\cellcolor{darkGray}{#1}}}
\newcommand{\second}[1]{{\cellcolor{mediumGray}{#1}}}
\newcommand{\third}[1]{{\cellcolor{lightGray}{#1}}}

\newcommand{\taskonomyscene}[1] {
    \includegraphics[height=\height]{#1/image.png} &
    \includegraphics[height=\height]{#1/depth.png} &
    \includegraphics[height=\height]{#1/pov-ground_truth.png} &
    \includegraphics[height=\height]{#1/pov-gecco.png} &
    \includegraphics[height=\height]{#1/pov-visualpriors.png} &
    \includegraphics[height=\height]{#1/bev-ground_truth.png} &
    \includegraphics[height=\height]{#1/bev-gecco.png} &
    \includegraphics[height=\height]{#1/bev-visualpriors.png}
}

\def \customparskip {.5em}
\renewcommand{\paragraph}[1]{\vspace{\customparskip}\noindent\textbf{#1}}
\newcommand{\para}[1]{\paragraph{{\bf #1.\!}}}

\def\ie{\emph{i.e}\onedot}

\usepackage[pagebackref=true,breaklinks=true,colorlinks=true,bookmarks=false,urlcolor=NavyBlue,citecolor=ForestGreen]{hyperref}

\iccvfinalcopy %

\def\iccvPaperID{2739} %

\ificcvfinal\fi

\begin{document}

\definecolor{backgroundpagecolor}{rgb}{0.9,0.9,0.9}

\setlength{\belowdisplayskip}{2pt}
\setlength{\belowdisplayshortskip}{2pt}
\setlength{\abovedisplayskip}{2pt}
\setlength{\abovedisplayshortskip}{2pt}

\newcommand{\highlight}[1]{{\underline{#1}}}
\title{\ours%
:
\highlight{Ge}ometri\highlight{c}ally-\highlight{Co}nditioned Point Diffusion Models
}

\author{%
    Michał J. Tyszkiewicz$^1$ \qquad  Pascal Fua$^1$ \qquad Eduard Trulls$^2$ \\
    $^1$École Polytechnique Fédérale de Lausanne (EPFL) \qquad $^2$Google Research, Zurich \\
    {\small \texttt{michal.tyszkiewicz@epfl.ch}} \quad {\small \texttt{pascal.fua@epfl.ch}} \quad {\small \texttt{trulls@google.com}} \\
}

\maketitle
\ificcvfinal\thispagestyle{empty}\fi

\begin{abstract}
Diffusion models generating images conditionally on text, such as Dall-E 2 \cite{ramesh2022dalle2} and Stable Diffusion\cite{rombach2022stablediffusion}, have recently made a splash far beyond the computer vision community. Here, we tackle the related problem of generating point clouds, both unconditionally, and conditionally with images. For the latter, we introduce a novel geometrically-motivated conditioning scheme based on projecting sparse image features into the point cloud and attaching them to each individual point, at every step in the denoising process. This approach improves geometric consistency and yields greater fidelity than current methods relying on unstructured, global latent codes.
Additionally, we show how to apply recent continuous-time diffusion schemes \cite{song2021scorebased,karras2022elucidating}. Our method performs on par or above the state of art on conditional and unconditional experiments on synthetic data, while being faster, lighter, and delivering tractable likelihoods. We show it can also scale to diverse indoors scenes.
\end{abstract}

\section{Introduction}
\label{sec:intro}

\renewcommand{\height}{1.42cm}
\begin{figure}
\centering
    \includegraphics[width=\linewidth]{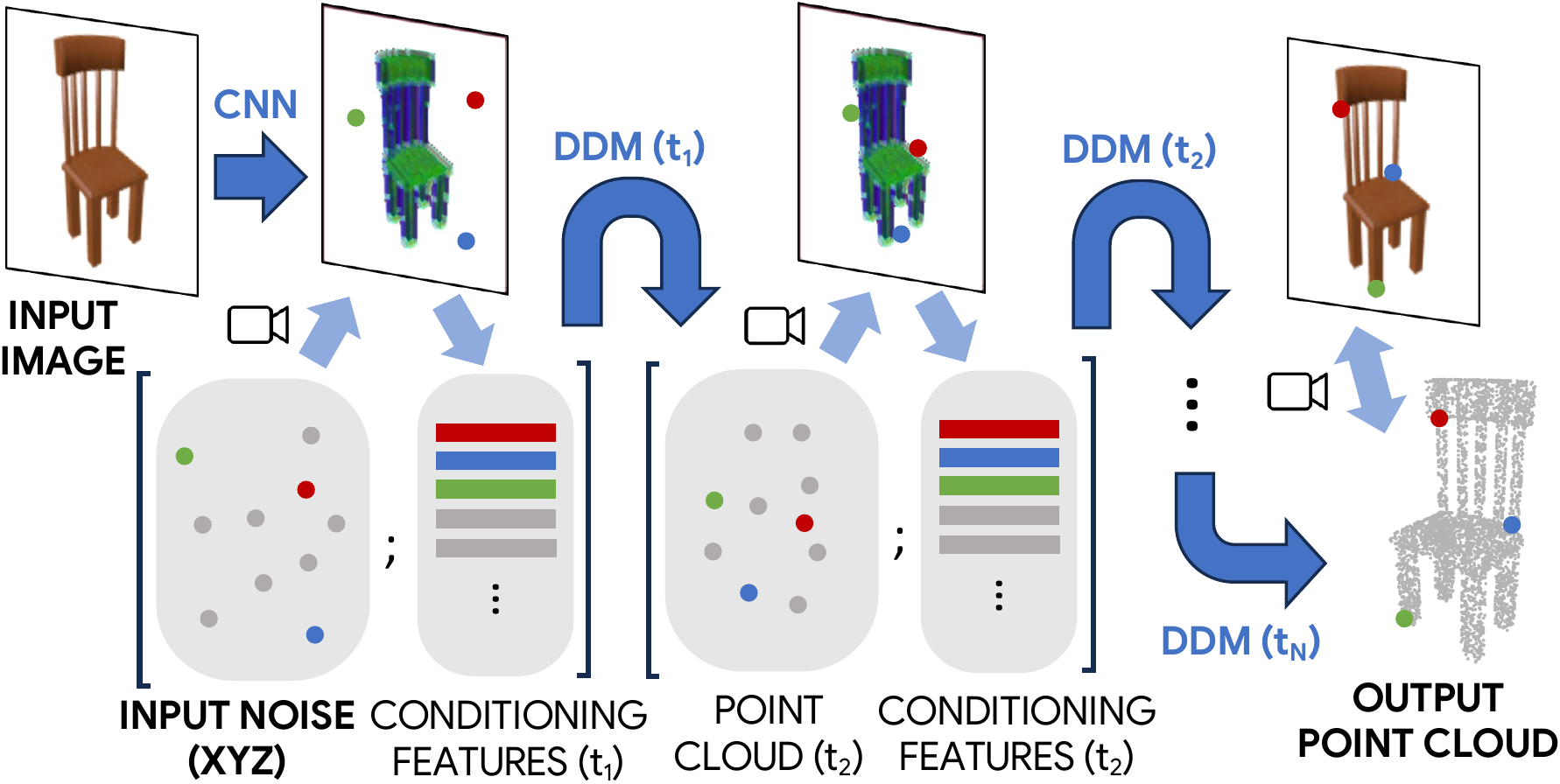}
    \caption{
    Our generative approach is based on denoising diffusion models (DDMs) and can be conditioned on images. At each denoising step we project the point cloud to the image, sample sparse features, and concatenate them to the locations, thus guiding the denoising process and yielding point clouds consistent with the images.}
\label{fig:diagram}
\end{figure}

Given the popularity of depth sensors and laser scanners, point clouds have become ubiquitous, with applications to robotics, autonomous driving, and augmented reality. Furthermore, they do not suffer from the precision/complexity trade-off inherent to voxel grids,
and scale better and more generally than graph-based representations such as meshes.
As a result, they have been extensively used for analytical tasks such as classification~\cite{qi2017pointnet,qi2017pointnet++,sun2020acne,zhao2021point,qian2022pointnext} and segmentation \cite{qi2017pointnet,qi2017pointnet++,hu2020randla,sun2020acne,genova2021learning,zhao2021point,zhu2021cylindrical}. 
Recent work has turned to point cloud synthesis and its many applications to 3D content creation. However, this remains an emerging field and state of the art methods \cite{yang2019pointflow,cai2020learning,kim2020softflow,luo2021diffusion,zhou2021pvd,zeng2022lion} operate on small datasets that feature only a handful of object types \cite{chang2015shapenet}.
More importantly, the generated shapes are typically not anchored in any prior and are thus difficult to control.

We present a novel approach that can mitigate these issues, taking our inspiration from generative methods that perturb samples with a diffusion process~\cite{sohl2015deep,karras2022elucidating} and denoise them with a deep network, which can later be used to synthesize new samples by iteratively denoising a signal.
Specifically, most denoising-based approaches generate novel samples from pure noise. But to mirror the success of text-based image synthesis \cite{ramesh2022dalle2,rombach2022stablediffusion,saharia2022photorealistic}, a generative approach must be able to not only produce samples of sufficient quality and diversity, but also ground them in contextual information.
Applying this generic idea to point clouds is, however, not straightforward.
We show how to achieve this by conditioning the network with sparse image features.

Unlike previous works relying on unstructured, global embeddings, we do so in a geometrically-principled way, by projecting the point cloud into an image, sampling sparse features at those locations, and feeding them to the network along with the point location, at each denoising step, as illustrated in Fig.~\ref{fig:diagram}. This allows us to render 3D objects geometrically and semantically consistent with the image content, while controlling the viewpoint. Unlike \mt{regression} models, such as monocular depth, our method can generate plausible hypotheses for occluded regions.
This work is thus a first step towards unlocking the applicability of denoising diffusion models to practical scenarios such as 3D content creation, generating priors for automotive or robotics applications, and single-view 3D reconstruction.

In short, we propose a novel generative point cloud model and show how to condition it on images.
Our main contributions are:
\begin{enumerate*}[label=(\roman*)]
    \item We propose a framework composed of a permutation-equivariant Set Transformer \cite{lee2019set} trained with a continuous-time diffusion scheme, which performs on par with the state of the art on {\it unconditional synthesis} while running 10x faster and delivering exact probabilities.
    \item We augment it with {\it geometrically-principled conditioning} to generate point clouds from images, yielding better reconstructions than with unstructured global embeddings, with state of the art performance.
    \item We bring denoising diffusion models for point cloud synthesis to the real world by applying our method to the Taskonomy dataset \cite{zamir2018taskonomy}.
\end{enumerate*}

\section{Related work}
\label{sec:related-work}

\para{Denoising diffusion models}
Denoising diffusion models \cite{sohl2015deep,ho2020denoising} are trained to denoise data perturbed by Gaussian noise. This process is applied iteratively during inference, and models are able to generate high-quality samples mirroring the distribution of the training data from random noise, optionally with a conditioning signal.
They have shown great success synthesizing images from text \cite{ramesh2022dalle2,rombach2022stablediffusion,saharia2022photorealistic}, speech \cite{chen2020wavegrad,kong2020diffwave,popov2021grad,chen2021wavegrad}, 3D objects \cite{jain2022zero,poole2022dreamfusion}, and recently point clouds \cite{luo2021diffusion,zhou2021pvd,zeng2022lion}.
Diffusion models can be applied discretely, with a Markov chain \cite{sohl2015deep,ho2020denoising}, or continuously with stochastic differential equations \cite{kingma2021variational,song2021scorebased}.
We use a continuous formulation first proposed in \cite{song2021scorebased}, specifically an extension proposed in \cite{karras2022elucidating}.

\renewcommand{\height}{2cm}
{\renewcommand{\arraystretch}{0.6}
\begin{figure}
\centering
\setlength{\tabcolsep}{1pt}
\begin{tabular}{@{}cccc@{}}
\includegraphics[height=\height]{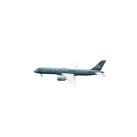} &
\includegraphics[height=\height]{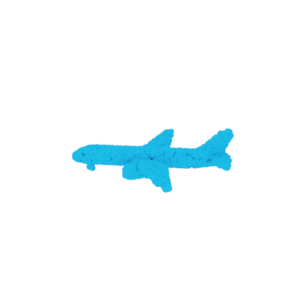} &
\includegraphics[height=\height]{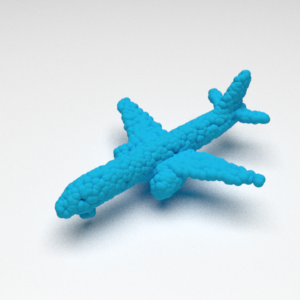} &
\includegraphics[height=\height]{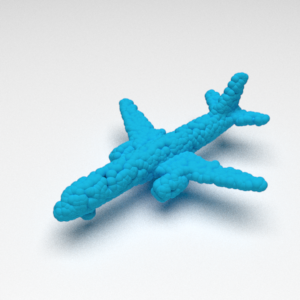} \\
\includegraphics[height=\height]{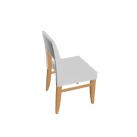} &
\includegraphics[height=\height]{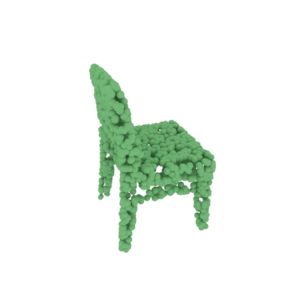} &
\includegraphics[height=\height]{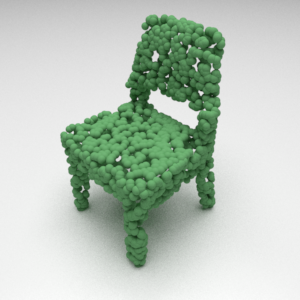} &
\includegraphics[height=\height]{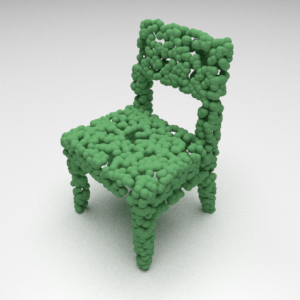} \\
\includegraphics[height=\height]{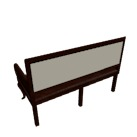} &
\includegraphics[height=\height]{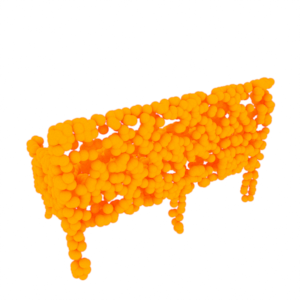} &
\includegraphics[height=\height]{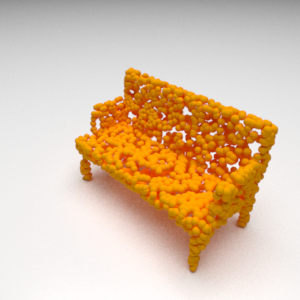} &
\includegraphics[height=\height]{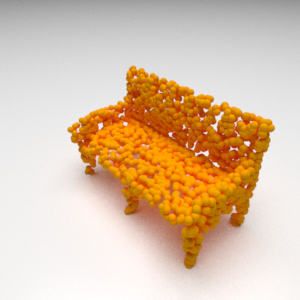} \\
\includegraphics[height=\height]{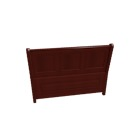} &
\includegraphics[height=\height]{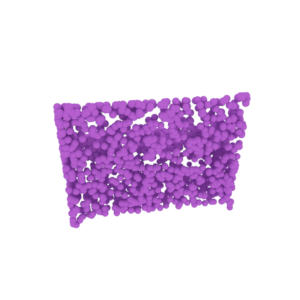} &
\includegraphics[height=\height]{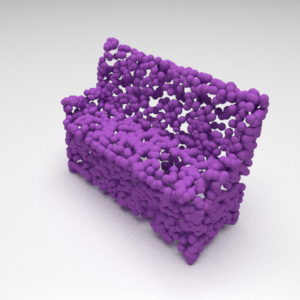} &
\includegraphics[height=\height]{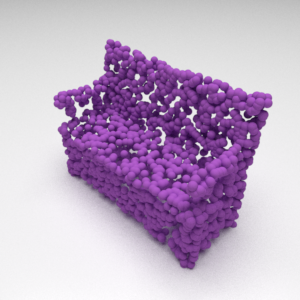} \\
\includegraphics[height=\height]{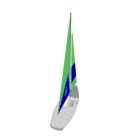} &
\includegraphics[height=\height]{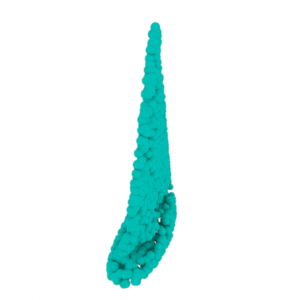} &
\includegraphics[height=\height]{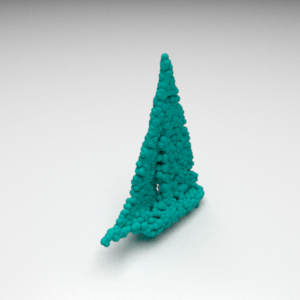} &
\includegraphics[height=\height]{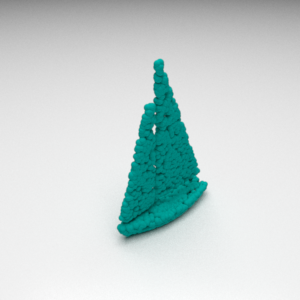} \\
\multirow{2}{*}{{\footnotesize (a) Source image}} & {\footnotesize (b) \ours{}} & {\footnotesize (c) \ours{}} & {\footnotesize (d) Ground Truth} \\
& {\footnotesize (source POV)} & {\footnotesize (canonical POV)} & {\footnotesize (canonical POV)} \\
\end{tabular}
\vspace{-1mm}
\caption{{\bf Image-conditioned generation.} We condition denoising diffusion models with images (a) in a geometrically-principled manner. Our model can reconstruct the shapes accurately from that view (b), and also generate plausible completions for regions not visible in the image (c, d), even under drastic occlusions (rows 3-5).}
\vspace{-4mm}
\label{fig:image-conditional}
\end{figure}

\para{Generative point clouds models}
Point cloud synthesis has been tackled with a wide array of techniques, including Variational Auto-Encoders (VAEs) \cite{kim2021setvae}, Generative Adversarial Networks (GANs) \cite{wu2016learning,achlioptas2018learning,li2018point,chen2019learning,shu20193d,hui2020progressive,kim2021setvae,li2021sp}, and autoregressive models \cite{sun2020pointgrow}.
Set-VAE \cite{kim2021setvae} proposed an attention-based hierarchical VAE applicable to sets, such as point clouds.
Achlioptas \etal \cite{achlioptas2018learning} introduced l-GAN, operating over latents encoding shape, and r-GAN, directly on point clouds.
SP-GAN \cite{li2021sp} guides the generator with a global, uniformly distributed spherical prior and a local, random latent code to disentangle global and local shape.
PointGrow \cite{sun2020pointgrow} relies on an autoregressive model that samples each point conditionally on previously-generated points.
\shapegf{} \cite{cai2020learning} learns distributions over gradient fields, moving randomly sampled points to high-density areas such as surfaces.
Xie \etal \cite{xie2021generative} formulate a permutation-invariant energy-based model with a PointNet.
Of more direct importance to us are two other families, discussed separately: those based on normalizing flows (NF) and denoising diffusion models (DDM).

\para{NFs for point cloud synthesis} Normalizing flows make powerful generative models and have been applied to point cloud synthesis \cite{yang2019pointflow,kim2020softflow,pumarola2020cflow,klokov2020discrete,postels2021go}.
\pointflow{} \cite{yang2019pointflow} broke ground by proposing a framework
dividing the problem into two stages. First, a latent code responsible for the shape of the object, $s \sim P_{\theta_s}$, is sampled. Second, individual points $p_i$ are sampled i.i.d.~conditionally on $s$, meaning that a cloud $\set{p_i}$ is sampled with probability
\begin{equation}
    P(\set{p_i}) = \int_{s \in \mathcal{S}} P_{\theta_s}(s) \sum_{i=0}^{n} P_{\theta_p}(p_i | s).
\end{equation}\label{eq:latent-shape}
A downside of this formulation is the intractability of the probability computation, requiring an integral over all~$s$: they thus train with an ELBO loss.
In contrast, \cflow{} \cite{pumarola2020cflow} models the point cloud jointly with a NF. This avoids the intractable probability, but they encounter difficulty in defining invertible layers which respect the permutation equivariance of point clouds. Their solution is to canonicalize the order of points with space-filling curves, which makes the approach complex. An alternative would be to use continuous-time NFs \cite{grathwohl2018ffjord}.
Unfortunately, powerful continuous-time models are known to be slow and expensive to train \cite{kelly2020learning}: as training progresses and the dynamics of the ODE become more complex, the cost of solving it with the precision necessary for stable backpropagation becomes impractically large.

\para{DDMs for point cloud synthesis}
This family, which our approach belongs to, has seen significant developments over the past year \cite{luo2021diffusion,zhou2021pvd,zeng2022lion}.
\diffpointflow{} \cite{luo2021diffusion} revisited the \pointflow{} \cite{yang2019pointflow} formulation (Eq.~\ref{eq:latent-shape}), replacing the normalizing flow $P(p_i|s)$ with a diffusion model. It relies on shape latents to parameterize the prior distribution with NFs, and uses them to condition a discrete DDM.
It splits the loss into two additive terms, which requires tuning hyperparmeters -- our approach is simpler.
\pvd{} \cite{zhou2021pvd} trains a discrete DDM directly on point clouds (without shape latents) using a point-voxel network (PVCNN) that enables 3D convolutions \cite{liu2019point}. It can optionally take in depth images as input to perform shape completion on occluded regions. This is achieved by freezing a set of points, extracted from the depth map, and optimizing over a set of `free' points -- we do geometrically-principled conditioning with RGB images instead, which is more widely applicable.
Moreover, the authors argue that conventional permutation-equivariant architectures operating on pure point representations such as PointNet++ \cite{qi2017pointnet++} are difficult to apply to diffusion models -- we show we can achieve similar performance with a very simple architecture \cite{lee2019set}.
In a different direction, PDR \cite{lyu2021conditional} proposes a dual-network approach to shape completion based on DDMs.

More recently, \lion{} \cite{zeng2022lion} proposed another two-stage approach. First, VAEs are used to obtain latent representations of both global shape, and points. Second, a DDM is trained to model those latent spaces.
They use a continuous diffusion model, like we do, but their reliance on shape latents means that computing exact probabilities is not tractable. Like \pvd{}, \lion{} uses PVCNN \cite{liu2019point} for the encoder, decoder, and diffusion models. Finally, it may condition samples with different signals, such as images or text embeddings, by conditioning the shape latent with adaptive Group Normalization in the PVCNN layers.
In contrast, we use a convolutional backbone to extract image features at the locations 3D points project to, concatenate them to the positional features, and feed them to the network.

\para{Other single-view reconstruction approaches}
Most of the generative point cloud methods our approach belongs to tackle only the unconditional problem. There is a wide array of relevant works on shape synthesis from single images, including regression and generative models, and using different representations such as voxels or meshes.
3D-R$^2$N$^2$ \cite{choy20163d} maps multiview images to occupancy grids with recurrent networks, but its resolution is limited due to using voxel grids.
Pixel2Mesh \cite{wang2018pixel2mesh} produces meshes from images by progressively deforming an ellipsoid using intermediate features extracted from the image.
AtlasNet \cite{groueix2018papier}, also mesh-based, generates 3D shapes by mapping multiple squares to shape surfaces.
Chen \etal \cite{chen2021unsupervised} learn point clouds from images by supervising their projections onto the image plane with samples from ground-truth silhouettes.
Pix2Point \cite{leroy2021pix2point} uses a 2D-3D hybrid network with an optimal transport loss to reconstruct point clouds from outdoor images.
PSGN \cite{fan2017point} combines an image and a random vector with a feedforward network to turn them into a point cloud, supervising with a permutation-equivariant loss.
OccNet \cite{onet} tackles 3D reconstruction as the decision boundary of a learned occupancy classifier, which unlike voxel-based methods can be evaluated at arbitrary resolutions. Finally, there is of course a large body of work on monocular depth estimation: we refer the reader to \cite{ming2021deep}.
\section{Method}
In order to follow the framework of \cite{karras2022elucidating,song2021scorebased}, we need to design a network $s_{\theta}$ to approximate the score $s_{\theta}(\mathbf{p}, t, c) \approx \nabla_{\mathbf{p}}\log p_t(\mathbf{p} | c)$, where $c$ is an optional conditioning signal. We treat the point set $\set{p_i}, p_i \in \mathbb{R}^D, i=1,\dots,N$ as a vector $\mathbf{p} \in \mathbb{R}^{N \times D}$. Since our network $s_{\theta}$ is permutation-equivariant, working with $\mathbf{p}$ is sound. We present experiments with $D=3$, but the approach is general.

\subsection{Score network}
\label{sec:score}

Our score network $s_{\theta}(\mathbf{p}, t, c)$ (where dependency on $c$ is optional)
is inspired by the Set Transformer \cite{lee2019set}, which we choose as a powerful permutation-invariant architecture specifically designed for unordered inputs, such as point clouds. It treats each point as a token, but due to the quadratic scaling of self-attention, it instead uses cross-attention with a number of \textit{inducers}, whose initial values are learned. Compared to the original Set Transformer, in each layer we use an extra shallow MLP on the inducers and not just on the tokens. Specifically to the task of diffusion, we ``inject'' the noise parameter $t$ through the bias and scale of the Group Normalization \cite{wu2018group} layers in the network, similarly to \cite{zeng2022lion}. We also follow the recent approach of \cite{ramasinghe2022beyond} and use Gaussian activations, allowing us to use a simple linear projection of input coordinates $\mathbb{R}^3 \rightarrow \mathbb{R}^{\dimnn}$, where $\dimnn$ is the dimensionality of the input to the Set Transformer. We found this approach substantially better than the more common Fourier feature embedding.

\subsection{Image-based conditioning}
\label{sec:method-conditioning}

\renewcommand{\width}{1mm}
\renewcommand{\height}{1.8cm}

{\renewcommand{\arraystretch}{0.6}
\begin{figure*}
\centering
\setlength{\tabcolsep}{1pt}
\begin{tabular}{@{}ccccccccccc@{}}
\includegraphics[height=\height]{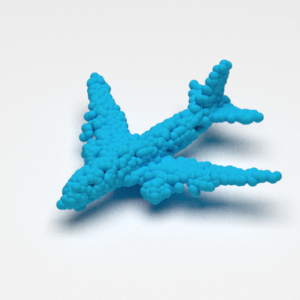} &
\includegraphics[height=\height]{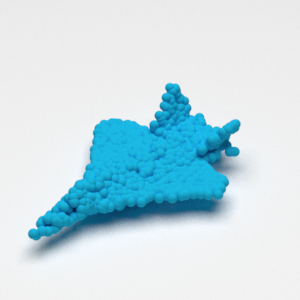} &
\hspace{\width} &
\includegraphics[height=\height]{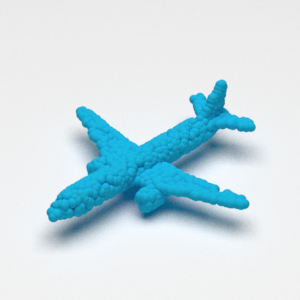} &
\includegraphics[height=\height]{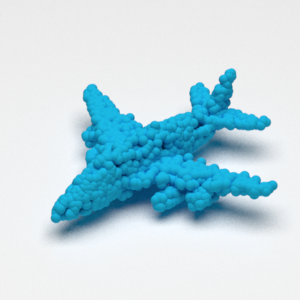} &
\includegraphics[height=\height]{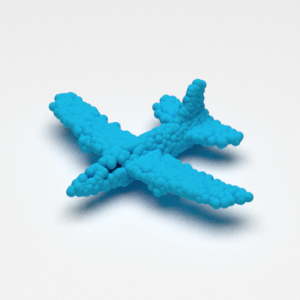} &
\includegraphics[height=\height]{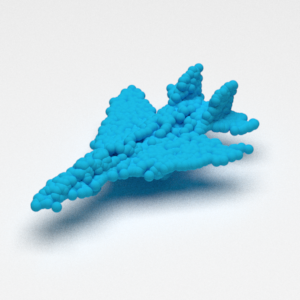} &
\includegraphics[height=\height]{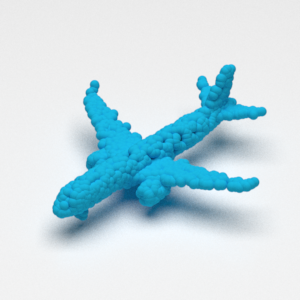} &
\hspace{\width} &
\includegraphics[height=\height]{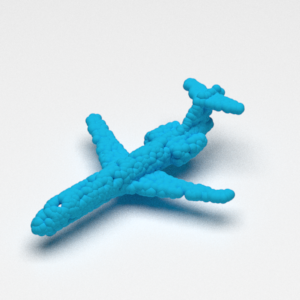} &
\includegraphics[height=\height]{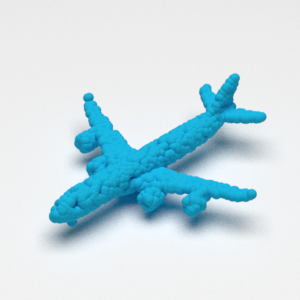} \\
\includegraphics[height=\height]{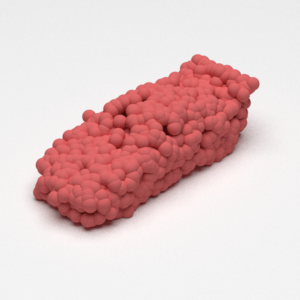} &
\includegraphics[height=\height]{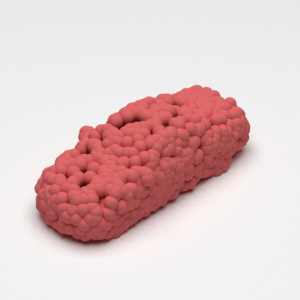} &
\hspace{\width} &
\includegraphics[height=\height]{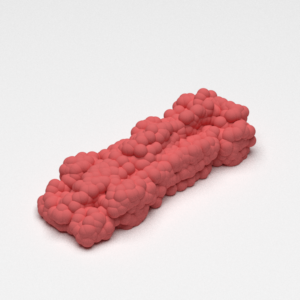} &
\includegraphics[height=\height]{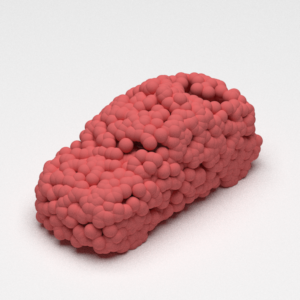} &
\includegraphics[height=\height]{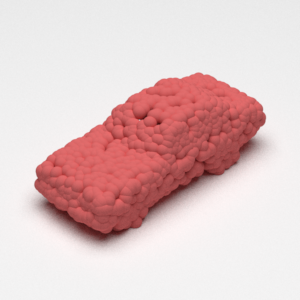} &
\includegraphics[height=\height]{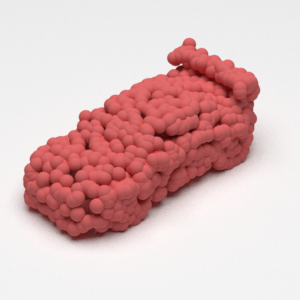} &
\includegraphics[height=\height]{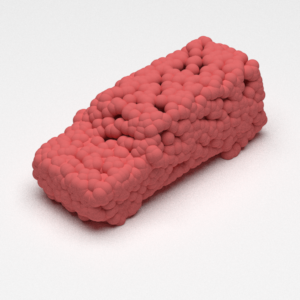} &
\hspace{\width} &
\includegraphics[height=\height]{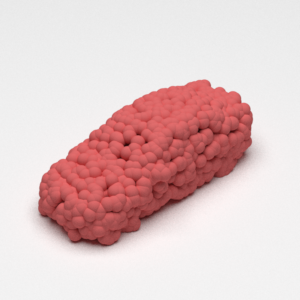} &
\includegraphics[height=\height]{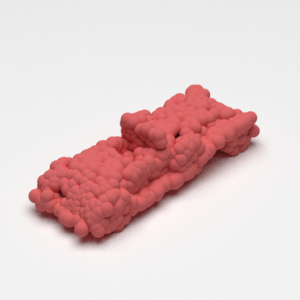} \\
\includegraphics[height=\height]{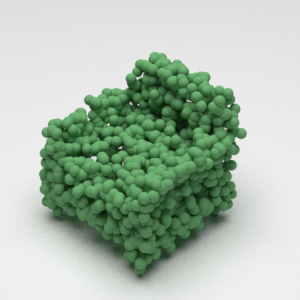} &
\includegraphics[height=\height]{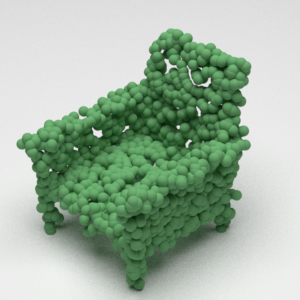} &
\hspace{\width} &
\includegraphics[height=\height]{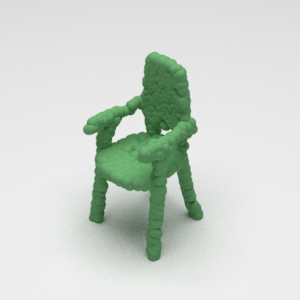} &
\includegraphics[height=\height]{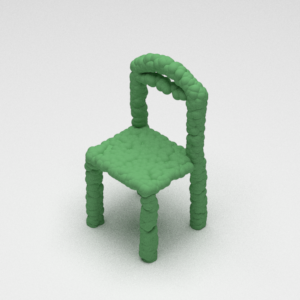} &
\includegraphics[height=\height]{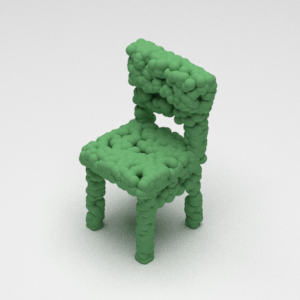} &
\includegraphics[height=\height]{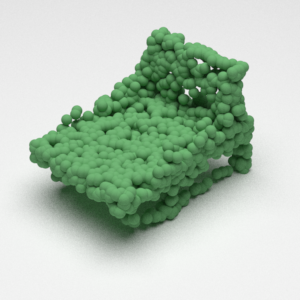} &
\includegraphics[height=\height]{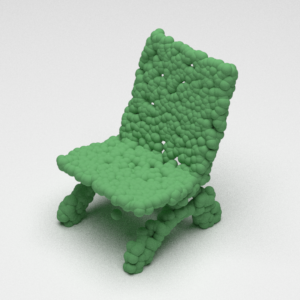} &
\hspace{\width} &
\includegraphics[height=\height]{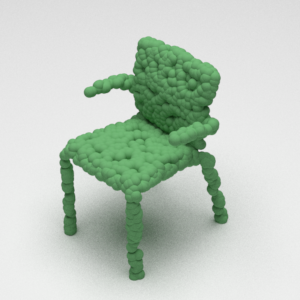} &
\includegraphics[height=\height]{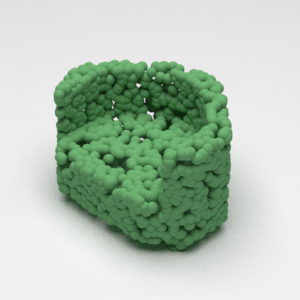} \\
{\footnotesize \diffpointflow~\cite{luo2021diffusion}} & {\footnotesize\shapegf~\cite{cai2020learning}} & & \multicolumn{5}{c}{{\footnotesize \ours{} (ours)}} & & \multicolumn{2}{c}{{\footnotesize Ground Truth}} \\
\end{tabular}
\vspace{-2mm}
\caption{{\bf Unconditional point cloud synthesis on ShapeNet.} All examples contain 2048 points.}
\label{fig:unconditional}
\vspace{-1mm}
\end{figure*}}

We undertake the goal of geometrically-principled point cloud generation conditioned on images. Specifically, we wish to: (a) generate point clouds in the reference frame of the camera; (b) accurately reconstruct the visible part of the object; and (c) build plausible hypotheses for occluded or ambiguous regions in a generative fashion.
Note that the last property sets our approach apart from most supervised approaches, such as monocular depth, which treat the problem point/pixel-wise and largely ignore the different modes of the posterior. Property (a) differs from most prior work on conditional generative models for point clouds, which usually condense the image into an unstructured, global embedding, losing much of the geometric detail. We achieve our goals through \textit{camera frustum reparameterization} of the point cloud and \textit{point-wise projective conditioning}.

\paragraph{Projective conditioning.} In order to provide the network with a highly accurate conditioning signal we need to maintain a geometrical interpretation of the model. We do so by using a ConvNeXt backbone \cite{liu2022convnet} to extract multi-scale image features. At any time $t$ in the diffusion process, we take the point cloud $\set{p_i}_t$, project the points onto the image plane, look up the image features corresponding to each projection using bilinear interpolation, and concatenate them to the point's location before feeding them to the transformer. 
This way each point knows the individual image properties {\em at its location}. We use this approach, which has a small computational overhead, to condition the reverse diffusion process {\em at every step}, as shown in Fig.~\ref{fig:diagram}.
We concatenate the point location and its associated feature prior to the MLP that projects the features to $\dimnn$ dimensions, and use the exact same transformer architecture for conditional and unconditional settings, making this the only difference between $s_\theta(\mathbf{p}, t)$ and $s_\theta(\mathbf{p}, t, c)$. For points outside of image bounds the bilinear look-up simply returns zeros.

\paragraph{Camera frustum reparameterization.} Current diffusion models are constrained to well-centered small objects and thus work directly in $\mathbb{R}^n$, but in Sec. \ref{sec:experiments-taskonomy} we wish to model points contained in the camera's viewing frustum only. Our solution is to \textit{reparameterize} the coordinates. Since the projection of each point on the image plane $(p_h, p_w) \in [0, 1]^2$, and its depth $p_d \in \mathbb{R}^+$, we can biject it to
\begin{equation}
    (u, v, l) = \left(S^{-1}(p_h), S^{-1}(p_w), \log(p_d) \right) \in \mathbb{R}^3,
\end{equation}
where $S^{-1}$ is the inverse sigmoid function. We use standard diffusion in $(u, v, l)$ and map points back to
$(x, y, z)$
at the end.
We do not use any reparameterization for ShapeNet.

\para{Implementation details}
Our Set Transformer-based network has 6 layers and takes inputs of dimensionality $\dimnn=384$. For unconditional models, we simply project the 3D point location to $\dimnn$. For image-conditioned models, we extract multi-scale ConvNeXt features of sizes 96, 192, 384 (total: 672), concatenate them to the point's location, and project them to $\dimnn$.
We train our models with 1024 or 2048 points, subsampling datasets which contain more points per example, for data augmentation.
We follow \cite{mingpt} in initializing transformer skip-connections with small weights and optimise the network (including the ConvNeXt) with AdaBelief \cite{zhuang2020adabelief} with learning rate $2 \cdot 10^{-4}$, for a number of steps depending on the dataset (see appendix). For each dataset we scale the data globally to zero mean, unit variance (which we undo at inference time) and pick $\sigma_\mathrm{max}$ by estimating the maximum pairwise distance between training examples. We train using the preconditioning and loss formulation of \cite{karras2022elucidating}, with the main departure in that we sample $\sigma$ values log-uniformly over $(10^{-4}, \sigmamax)$ instead of log-normally.
As in \cite{song2021scorebased}, we found it beneficial to apply an exponential moving average to the weights of our model: we use a rate of 0.999.
We implement our software with JAX \cite{jax2018github} and Equinox \cite{kidger2021equinox} and use Diffrax \cite{kidger2021diffrax} for ODE solving.
For inference we use the 2nd-order stochastic sampler of \cite{karras2022elucidating} (later referred to as `SDE') or the probability flow of \cite{song2021scorebased} (later: `ODE'),  with 128 steps. Please refer to Sec.~\ref{sec:experiments-ablation} for an ablation study, and to Table~\ref{tab:runtime} for computational details: our approach is both {\em faster} and {\em lighter} than comparable methods.
\et{Code and models are available.}
{\renewcommand{\arraystretch}{0.9}
\setlength{\tabcolsep}{3pt}
\begin{table}
\footnotesize
\centering
\begin{tabular}{@{}cccccccc@{}}
\toprule
& & \multicolumn{2}{c}{MMD$\downarrow$} & \multicolumn{2}{c}{COV$\uparrow$ (\%)} & \multicolumn{2}{c}{1-NNA$\downarrow$ (\%)} \\
& Model & CD & EMD & CD & EMD & CD & EMD \\
\midrule
\multirow{14}{*}{\rotatebox{90}{\textsc{airplane}}} & Oracle & 0.214 & 0.369 & 46.17 & 49.88 & 64.44 & 63.58 \\
\cmidrule(l){2-8}
& r-GAN~\cite{achlioptas2018learning} & 0.447 & 2.309 & 30.12 & 14.32 & 98.40 & 96.79 \\
& l-GAN-CD~\cite{achlioptas2018learning} & 0.340 & 0.583 & 38.52 & 21.23 & 87.30 & 93.95 \\
& l-GAN-EMD~\cite{achlioptas2018learning} & 0.397 & 0.417 & 38.27 & 38.52 & 89.49 & 76.91 \\
& PointFlow~\cite{yang2019pointflow} & 0.224 & 0.390 & 47.90 & 46.41 & 75.68 & 70.74 \\
& SoftFlow~\cite{kim2020softflow} & 0.231 & 0.375 & 46.91 & 47.90 & 76.05 & 65.80 \\
& SetVAE~\cite{kim2021setvae} & \first{0.200} & \first{0.367} & 43.70 & 48.40 & 76.54 & 67.65 \\
& DPF-Net~\cite{klokov2020discrete} & 0.264 & 0.409 & 46.17 & \third{48.89} & 75.18 & 65.55 \\
& DPM~\cite{luo2021diffusion} & \second{0.213} & 0.572 & \second{48.64} & 33.83 & 76.42 & 86.91 \\
& PVD~\cite{zhou2021pvd} & 0.224 & \third{0.370} & \first{48.88} & \first{52.09} & \third{73.82} & \third{64.81} \\
& LION~\cite{zeng2022lion} & \third{0.219} & 0.372 & 47.16 & \second{49.63} & \first{67.41} & \first{61.23} \\
\cmidrule(l){2-8}
& \ours{} (ours) & 0.245 & \second{0.368} & \third{48.15} & 48.40 & \second{72.10} & \second{62.96} \\
\midrule
\multirow{14}{*}{\rotatebox{90}{\textsc{chair}}} & Oracle & 2.618 & 1.555 & 53.02 & 51.21 & 51.28 & 54.76 \\
\cmidrule(l){2-8}
& r-GAN~\cite{achlioptas2018learning} & 5.151 & 8.312 & 24.27 & 15.13 & 83.69 & 99.70 \\
& l-GAN-CD~\cite{achlioptas2018learning} & 2.589 & 2.007 & 41.99 & 29.31 & 68.58 & 83.84 \\
& l-GAN-EMD~\cite{achlioptas2018learning} & 2.811 & 1.619 & 38.07 & 44.86 & 71.90 & 64.65 \\
& PointFlow~\cite{yang2019pointflow} & \second{2.409} & 1.595 & 42.90 & \third{50.00} & 62.84 & 60.57 \\
& SoftFlow~\cite{kim2020softflow} & \third{2.528} & 1.682 & 41.39 & 47.43 & 59.21 & 60.05 \\
& SetVAE~\cite{kim2021setvae} & 2.545 & \third{1.585} & \third{46.83} & 44.26 & 58.84 & 60.57 \\
& DPF-Net~\cite{klokov2020discrete} & 2.536 & 1.632 & 44.71 & 48.79 & 62.00 & 58.53 \\
& DPM~\cite{luo2021diffusion} & \first{2.399} & 2.066 & 44.86 & 35.50 & 60.05 & 74.77 \\
& PVD~\cite{zhou2021pvd} & 2.622 & \second{1.556} & \first{49.84} & \second{50.60} & \second{56.26} & \second{53.32} \\
& LION~\cite{zeng2022lion} & 2.640 & \first{1.550} & \second{48.94} & \first{52.11} & \first{53.70} & \first{52.34} \\
\cmidrule(l){2-8}
& \ours{} (ours) & 2.793 & 1.601 & 46.68 & 49.40 & \third{56.57} & \third{54.68} \\
\midrule
\multirow{14}{*}{\rotatebox{90}{\textsc{car}}} & Oracle & 0.938 & 0.791 & 50.85 & 55.68 & 51.70 & 50.00 \\
\cmidrule(l){2-8}
& r-GAN~\cite{achlioptas2018learning} & 1.446 & 2.133 & 19.03 & 6.539 & 94.46 & 99.01 \\
& l-GAN-CD~\cite{achlioptas2018learning} & 1.532 & 1.226 & 38.92 & 23.58 & 66.49 & 88.78 \\
& l-GAN-EMD~\cite{achlioptas2018learning} & 1.408 & 0.899 & 37.78 & 45.17 & 71.16 & 66.19 \\
& PointFlow~\cite{yang2019pointflow} & \second{0.901} & 0.807 & 46.88 & 50.00 & 58.10 & 56.25 \\
& SoftFlow~\cite{kim2020softflow} & 1.187 & 0.859 & 42.90 & 44.60 & 64.77 & 60.09 \\
& SetVAE~\cite{kim2021setvae} & \first{0.882} & \first{0.733} & \third{49.15} & 46.59 & 59.94 & 59.94 \\
& DPF-Net~\cite{klokov2020discrete} & 1.129 & 0.853 & 45.74 & 49.43 & 62.35 & 54.48 \\
& DPM~\cite{luo2021diffusion} & \third{0.902} & 1.140 & 44.03 & 34.94 & 68.89 & 79.97 \\
& PVD~\cite{zhou2021pvd} & 1.077 & 0.794 & 41.19 & \third{50.56} & \second{54.55} & \third{53.83} \\
& LION~\cite{zeng2022lion} & 0.913 & \second{0.752} & \first{50.00} & \second{56.53} & \first{53.41} & \second{51.14} \\
\cmidrule(l){2-8}
& \ours{} (ours) & 1.044 & \third{0.769} & \first{50.00} & \first{56.82} & \third{56.82} & \first{49.15} \\
\bottomrule
\end{tabular}
\caption{{\bf Unconditional generation (global normalization).} Generation metrics on three ShapeNet categories. MMD-CD is multiplied by 10$^3$, and MMD-EMD by 10$^2$. The top 3 are highlighted \et{in gray (darker is better).} %
}
\vspace{-4mm}
\label{tab:shapenet-unconditional-global}
\end{table}}
{\renewcommand{\arraystretch}{0.9}
\setlength{\tabcolsep}{3pt}
\begin{table}
\footnotesize
\centering
\begin{tabular}{@{}cccccccc@{}}
\toprule
& & \multicolumn{2}{c}{MMD$\downarrow$} & \multicolumn{2}{c}{COV$\uparrow$ (\%)} & \multicolumn{2}{c}{1-NNA$\downarrow$ (\%)} \\
& Model & CD & EMD & CD & EMD & CD & EMD \\
\midrule
\multirow{10}{*}{\rotatebox{90}{\textsc{airplane}}} & Oracle & 0.230 & 0.539 & 42.72 & 45.68 & 69.26 & 67.78 \\
\cmidrule(l){2-8}
& TreeGAN~\cite{shu20193d} & 0.558 & 1.460 & 31.85 & 17.78 & 97.53 & 99.88 \\
& ShapeGF~\cite{cai2020learning} & \first{0.313} & \third{0.636} & 45.19 & 40.25 & \third{81.23} & \third{80.86} \\
& SP-GAN~\cite{li2021sp} & 0.403 & 0.766 & 26.42 & 24.44 & 94.69 & 93.95 \\
& PDGN~\cite{hui2020progressive} & 0.409 & 0.701 & 38.77 & \third{36.54} & 94.94 & 91.73 \\
& GCA~\cite{zhang2021learning} & 0.359 & 0.765 & 38.02 & 36.30 & 88.15 & 85.93 \\
& LION~\cite{zeng2022lion} & \third{0.356} & \second{0.593} & 42.96 & \second{47.90} & \second{76.30} & \first{67.04} \\
\cmidrule(l){2-8}
& GECCO (ours) & \second{0.354} & \first{0.572} & 44.20 & \first{50.12} & \first{76.17} & \second{68.89} \\
\midrule
\multirow{10}{*}{\rotatebox{90}{\textsc{chair}}} & Oracle & 3.864 & 2.302 & 49.7 & 42.11 & 55.14 & 54.76 \\
\cmidrule(l){2-8}
& TreeGAN~\cite{shu20193d} & 4.841 & 3.505 & 39.88 & 26.59 & 88.37 & 96.37 \\
& ShapeGF~\cite{cai2020learning} & \first{3.724} & \second{2.394} & \second{48.34} & 44.26 & \third{58.01} & \third{61.25} \\
& SP-GAN~\cite{li2021sp} & 4.208 & 2.620 & 40.03 & 32.93 & 72.58 & 83.69 \\
& PDGN~\cite{hui2020progressive} & 4.224 & 2.577 & 43.20 & 36.71 & 71.83 & 79.00 \\
& GCA~\cite{zhang2021learning} & 4.403 & 2.582 & 45.92 & \third{47.89} & 64.27 & 64.50 \\
& LION~\cite{zeng2022lion} & \second{3.846} & \first{2.309} & \third{46.37} & \second{50.15} & \second{56.50} & \first{53.85} \\
\cmidrule(l){2-8}
& GECCO (ours) & \third{4.119} & \third{2.410} & \first{48.64} & \first{52.42} & \first{55.36} & \second{56.80} \\
\midrule
\multirow{10}{*}{\rotatebox{90}{\textsc{car}}} & Oracle & 1.05 & 0.829 & 47.44 & 48.01 & 57.53 & 56.68 \\
\cmidrule(l){2-8}
& TreeGAN~\cite{shu20193d} & 1.142 & 1.063 & 40.06 & 31.53 & 89.77 & 94.89 \\
& ShapeGF~\cite{cai2020learning} & \first{1.020} & \third{0.824} & \second{44.03} & 47.16 & \third{61.79} & \third{57.24} \\
& SP-GAN~\cite{li2021sp} & 1.168 & 1.021 & 34.94 & 31.82 & 87.36 & 85.94 \\
& PDGN~\cite{hui2020progressive} & 1.184 & 1.063 & 31.25 & 25.00 & 89.35 & 87.22 \\
& GCA~\cite{zhang2021learning} & 1.074 & 0.867 & 42.05 & \third{48.58} & 70.45 & 64.20 \\
& LION~\cite{zeng2022lion} & \third{1.064} & \second{0.808} & \third{42.90} & \first{50.85} & \first{59.52} & \second{49.29} \\
\cmidrule(l){2-8}
& GECCO (ours) & \second{1.063} & \first{0.802} & \first{46.31} & \second{49.15} & \second{60.51} & \first{47.87} \\
\bottomrule
\end{tabular}
\caption{{\bf Unconditional generation (per-shape normalization).} Same as Table~\ref{tab:shapenet-unconditional-global}, with per-shape normalization.}
\label{tab:shapenet-unconditional-per-sample}
\vspace{-2mm}
\end{table}}
{\renewcommand{\arraystretch}{0.9}
\setlength{\tabcolsep}{4pt}
\begin{table*}
\small
\centering
\resizebox{\textwidth}{!}{%
\begin{tabular}{lcccccccccccccc}
\toprule
& airplane & bench & cabinet & car & chair & display & lamp & loudspeaker & rifle & sofa & table & telephone & vessel & {\bf average} \\
\midrule
3D-R$^2$N$^2$ \cite{choy20163d} & 0.227 & 0.194 & 0.217 & 0.213 & 0.270 & 0.314 & 0.778 & 0.318 & 0.183 & 0.229 & 0.239 & 0.195 & 0.238 & 0.278 \\
PSGN \cite{fan2017point} & 0.137 & 0.181 & 0.215 & 0.169 & 0.247 & 0.284 & 0.314 & 0.316 & 0.134 & 0.224 & 0.222 & 0.161 & 0.188 & 0.215 \\
Pixel2Mesh \cite{wang2018pixel2mesh} & 0.187 & 0.201 & 0.196 & 0.180 & 0.265 & 0.239 & 0.308 & 0.285 & 0.164 & 0.212 & 0.218 & 0.149 & 0.212 & 0.216 \\
AtlasNet \cite{groueix2018papier} & {\bf 0.104} & 0.138 & 0.175 & 0.141 & 0.209 & 0.198 & 0.305 & 0.245 & 0.115 & 0.177 & 0.190 & 0.128 & 0.151 & 0.175 \\
OccNet \cite{onet} & 0.140 & 0.157 & 0.156 & 0.153 & 0.209 & 0.260 & 0.394 & 0.269 & 0.142 & 0.185 & 0.176 & 0.129 & 0.200 & 0.198 \\
\ours{} & 0.106 & {\bf 0.097} & {\bf 0.110} & {\bf 0.103} & {\bf 0.142} & {\bf 0.138} & {\bf 0.186} & {\bf 0.158} & {\bf 0.097} & {\bf 0.123} & {\bf 0.107} & {\bf 0.090} & {\bf 0.132} & {\bf 0.122} \\
\midrule
OccNet \cite{onet} w/ ICP & 0.151 & 0.158 & 0.141 & 0.139 & 0.196 & 0.247 & 0.380 & 0.251 & 0.155 & 0.188 & 0.207 & 0.138 & 0.203 & 0.196 (+1.02\%) \\
\ours{} w/ ICP & {\bf 0.081} & {\bf 0.088} & {\bf 0.100} & {\bf 0.093} & {\bf 0.117} & {\bf 0.119} & {\bf 0.164} & {\bf 0.141} & {\bf 0.073} & {\bf 0.114} & {\bf 0.111} & {\bf 0.083} & {\bf 0.128} & {\bf 0.109} (+11.9\%) \\
\bottomrule
\end{tabular}}
\caption{{\bf Image-conditional generation on ShapeNet-Vol.} L1 chamfer distance between samples reconstructed from an image and the ground truth point clouds (lower is better), following \cite{onet}.
Qualitative results are available in Fig.~\ref{fig:image-conditional}.
}
\label{tab:image-conditional}
\vspace{-2mm}
\end{table*}}

\section{Experiments}

We use ShapeNet \cite{chang2015shapenet} to evaluate unconditional point cloud synthesis in \Section{experiments-shapenet-unconditional}, and with image conditioning in \Section{experiments-shapenet-conditional}. We then show that our method can translate to larger-scale, real data on the Taskonomy dataset \cite{zamir2018taskonomy} in Sec.~\ref{sec:experiments-taskonomy}, and ablate it and showcase some of its properties in Sec.~\ref{sec:experiments-ablation}. We render point clouds with Mitsuba 3~\cite{jakob2022mitsuba3}.

\subsection{Unconditional generation on ShapeNet}
\label{sec:experiments-shapenet-unconditional}

\para{Dataset} We evaluate our approach on the dataset most commonly used for generative shape modelling: ShapeNet \cite{chang2015shapenet}. We follow the methodology, splits, and metrics introduced by \pointflow{} \cite{yang2019pointflow}, which provides point clouds sampled from the original meshes, and train single-class models for three categories: airplane, chair, and car. 
In order to ensure reproducibility we compare against the results published in \cite{zeng2022lion}, the most recent and thorough.
While \pointflow{} normalizes the data globally to zero-mean per axis, and unit variance, others methods use per-shape normalization: we consider both. We use 2048 points for all methods.

\para{Metrics} We consider two distance metrics between point clouds: the chamfer distance (CD), which measures the average squared distance between each point in one set to its nearest neighbor on the other set; and the earth mover's distance (EMD)%
, which solves the optimal transport problem. Given point sets $\mathbf{p}=\{p_i\}$ and $\mathbf{q}=\{q_i\}$ with $N$ points each, and $\phi$ a bijection between them, they are defined as:
{\small
\begin{equation}
\hspace{-1mm}\text{CD}(\mathbf{p}, \mathbf{q}) = \frac{1}{N} \left[ \sum_{p \in \mathbf{p}} \min_{q \in \mathbf{q}} \| p - q \|^2_2 + \sum_{q \in \mathbf{q}} \min_{p \in \mathbf{p}} \| p - q \|^2_2 \right],
\label{eq:chamfer}
\end{equation}}
\begin{equation}
\text{EMD}(\mathbf{p}, \mathbf{q}) = \frac{1}{N} \min_{\phi: \mathbf{p} \rightarrow \mathbf{q}} \sum_{p \in \mathbf{p}} \| p - \phi(p) \|_2.
\label{eq:emd}
\end{equation}
Given these two similarity functions, we sample as many point clouds as there are in the reference set $\mathbf{S_r}$ to obtain a generated set $\mathbf{S_g}$ and compute three metrics between the ground truth and sampled collections.
To compute {\bf coverage (COV)} we find the nearest neighbor in the reference set for each point cloud in the generated set, and compute the fraction of shapes in the reference set that are matched to at least one shape in the reference set. It can capture mode collapse, but not the quality of the samples.
The {\bf minimum matching distance (MMD)} is a complementary metric that measures the average minimum distance from every sample in the reference set to every sample in the generated set. \pointflow{} \cite{yang2019pointflow} proposes an arguably better metric also used for GANs: {\bf 1-Nearest Neighbour Accuracy (1-NNA)}. It is defined as the accuracy of a leave-one-out classifier that assigns each sample in $\mathbf{S_r} \cup \mathbf{S_g}$ to the `class' (set) of its closest neighbor, other than itself. Note that a perfect oracle would score $\sim$50\%. As all three metrics rely on nearest neighbours, they can be computed with CD or EMD: we report both. Details are provided in the supplementary material.

\para{Results} Results are shown in Table~\ref{tab:shapenet-unconditional-global} for global normalization, and Table~\ref{tab:shapenet-unconditional-per-sample} for per-shape normalization. 1-NNA is the metric favored by most recent papers. Our method performs on par with \lion{}, the state of the art, on the first benchmark and slightly outperforms it on the second. In addition to the baselines, we consider an oracle that spits out samples from the training set instead of generating novel ones. Our method outperforms this oracle in 1-NNA-EMD for all three categories in Table~\ref{tab:shapenet-unconditional-global}, which suggests that the dataset is at the saturation point, or that the distance metrics fail to fully capture the quality of the samples. We show samples and ground truth examples in Fig.~\ref{fig:unconditional}.

\subsection{Conditional generation on ShapeNet-Vol}
\label{sec:experiments-shapenet-conditional}

\para{Dataset} None of the baselines used in the previous section are able to condition the generative process with images, with the exception of LION, which may optionally train DDMs conditioned with CLIP embeddings \cite{radford2021learning,sanghi2022clip} extracted from ShapeNet renders. While seemingly effective this approach is not geometrically principled, and the paper offers only qualitative examples. We thus turn to the ShapeNet-Vol benchmark, originally introduced by 3D-R$^2$N$^2$ \cite{choy20163d}, which provides rendered images and voxelized models for 13 ShapeNet categories: each shape is rendered from 24 viewpoints at 137$\times$137 pixels. We align the point clouds to the camera pose for each view and train our models with the conditioning scheme of \Section{method-conditioning}. Note that unlike the unconditional experiments in the previous section, here we train {\em a single model for all 13 categories}.

\para{Results} We follow the evaluation protocol for single-view reconstruction introduced by OccNet \cite{onet}. For mesh-based methods, such as OccNet, the benchmark samples 100k points from the mesh and computes the chamfer distance between the generated and ground truth point clouds as a quality metric -- including points occluded in the image. For point-based methods such as PSGN \cite{fan2017point}, the benchmark simply samples more points (no meshing). In order to reach the required number of points, we simply generate multiple samples for each image and concatenate them. Note that following \cite{onet}, we use the {\em L1 chamfer distance}, defined as in Eq.~\ref{eq:chamfer} but without squaring the norms. OccNet reconstructs the model in a canonical reference frame, which is also used for evaluation, but our method generates predictions from the point of view of the camera: we move our predictions to this canonical frame with the ground truth pose. We report results in Table~\ref{tab:image-conditional}\footnote{For OccNet we run the latest model available in their \href{https://github.com/autonomousvision/occupancy_networks}{repository}, improved from the original paper. For others we use the results from \cite{onet}.}. Our models outperform all mesh-based methods, including OccNet, and also PSGN. Additionally, we noticed that our generated point clouds were slightly misaligned, but not those from OccNet. We hypothesized this was due to OccNet generating samples in a canonical reference frame rather than the camera's point of view, which while advantageous here does lose generalization. We confirmed this by aligning the point clouds with ICP~\cite{besl1992icp, ravi2020pytorch3d} and re-computing the metric: our method improves by 12\% relative, compared to OccNet's 1\%. Note that unlike OccNet, our approach does not need normalized data in a canonical pose and it can deal with non-watertight meshes.
On the other hand, it does require known camera intrinsics.
We show qualitative examples in Fig.~\ref{fig:image-conditional}.

\renewcommand{\height}{2.11cm}
{\renewcommand{\arraystretch}{0.6}
\begin{figure*}
\centering
\setlength{\tabcolsep}{1pt}
\begin{tabular}{@{}cccccccc@{}}
\taskonomyscene{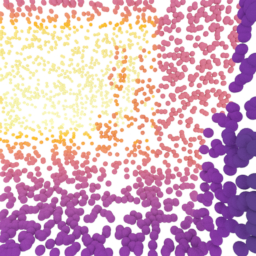}\\
\taskonomyscene{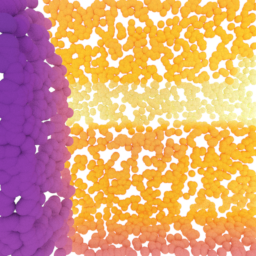}\\
\taskonomyscene{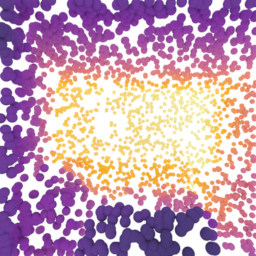}\\
{\footnotesize (a) Source image} & {\footnotesize (b) GT depth} & {\footnotesize (c) GT POV} & {\footnotesize (d) \ours{} POV} & {\footnotesize (e) \cite{sax2020visualpriors} POV} & {\footnotesize (f) GT BEV} & {\footnotesize (g) \ours{} BEV} & {\footnotesize (h) \cite{sax2020visualpriors} BEV} \\
\end{tabular}
\vspace{-2mm}
\caption{{\bf Qualitative examples on Taskonomy.} Color encodes depth. We show point clouds with 2048 points each, from the cameras's point of view, and from a bird's eye view. The red dot in BEV marks the location of the camera.}
\label{fig:omnidata}
\end{figure*}}

{\renewcommand{\arraystretch}{0.9}
\setlength{\tabcolsep}{4pt}
\begin{table}
\small
\centering
\begin{tabular}{llcc}
\toprule
Split & Model & Chamfer $\downarrow$ & Chamfer (ICP) $\downarrow$ \\
\midrule
\multirow{2}{*}{Test} & \ours &   0.661 / \textbf{0.502} & \textbf{0.444} / \textbf{0.242} \\
& Monocular depth~\cite{sax2020visualpriors} & \textbf{0.632} / 0.527 & 0.558 / 0.451 \\
\midrule
\multirow{2}{*}{Val} & \ours &   \textbf{0.541} / \textbf{0.427} & \textbf{0.361} / \textbf{0.222} \\
 & Monocular depth~\cite{sax2020visualpriors} & 0.567 / 0.490 & 0.497 / 0.418\\
\bottomrule
\end{tabular}

\caption{{\bf Evaluation on Taskonomy} with mean / median values.
Note that we do not use the validation set for early stopping, so `validation' and `test' both act as test sets. The difference in performance in `test' is due to out-of-distribution scenes in that subset (see appendix for details).}
\label{tab:taskonomy}
\vspace{-2mm}
\end{table}}

\subsection{Conditional generation on Taskonomy}
\label{sec:experiments-taskonomy}

We also wish to showcase how our method scales to real data, and past the object-centric nature of ShapeNet that most generative methods are limited to.
For this purpose we turn to Taskonomy~\cite{zamir2018taskonomy}, which contains a large dataset of scanned indoor scenes with high-quality depth maps, and convert them into point clouds by sampling and unprojecting 8192 points per image. We sample points inversely proportionally to pixel depth, to emulate per-surface-area densities. This yields a rich image-point cloud dataset.

\newcommand{\spacersize}{6}

{\renewcommand{\arraystretch}{0.8}
\setlength{\tabcolsep}{2pt}
\begin{table}
\small
\centering
\begin{tabular}{ccccccc}
\toprule
Conditioning & Sampler & $N_{\text{denoise}}$ & ICP & Subset & L1-CD$\downarrow$ & $\Delta\uparrow$ (\%) \\
\midrule
Global & ODE & 128 & \xmark & -- & 0.305 & -- \\
Projective & ODE & 128 & \xmark & -- & 0.286 & +6.6\% \\
Global & SDE & 128 & \xmark & -- & 0.302 & -- \\
Projective & SDE & 128 & \xmark & -- & 0.283 & +6.7\% \\
\midrule
Global & ODE & 128 & \cmark & -- & 0.280 & -- \\

Projective & ODE & 128 & \cmark & -- & 0.259 & +8.1\% \\
Global & SDE & 128 & \cmark & -- & 0.276 & -- \\
Projective & SDE & 128 & \cmark & -- & 0.257 & +7.4\% \\
\midrule
Projective & SDE & 8 & \xmark & -- & 0.615 & -117.3\% \\
Projective & SDE & 16 & \xmark & -- & 0.309 & -9.2\% \\
Projective & SDE & 32 & \xmark & -- & 0.286 & -1.1\% \\
Projective & SDE & 64 & \xmark & -- & 0.287 & -1.4\% \\
Projective & SDE & 128 & \xmark & -- & 0.283 & -- \\
\midrule
Projective & SDE & 8 & \cmark & -- & 0.353 & -37.4\% \\
Projective & SDE & 16 & \cmark & -- & 0.267 & -3.9\% \\
Projective & SDE & 32 & \cmark & -- & 0.258 & -0.4\% \\
Projective & SDE & 64 & \cmark & -- & 0.258 & -0.4\% \\
Projective & SDE & 128 & \cmark & -- & 0.257 & -- \\
\midrule
Global & SDE & 128 & \xmark & 50\% & 0.308 & -2.0\% \\
Global & SDE & 128 & \cmark & 50\% & 0.283 & -2.5\% \\
Projective & SDE & 128 & \xmark & 50\% & 0.289 & -2.1\% \\
Projective & SDE & 128 & \cmark & 50\% & 0.261 & -1.6\% \\
\bottomrule
\end{tabular}
\vspace{-2mm}
\caption{{\bf Ablation study on the OccNet benchmark.} We compare conditioning with global context vs projective lookups (sec.~\ref{sec:method-conditioning}), probability flow (`ODE') \cite{song2021scorebased} vs the stochastic solver (`SDE') of \cite{karras2022elucidating}, the number of solver iterations, and the impact of reducing the training data to 50\%. We conduct these experiments with 1024 points, for speed.
}
\label{tab:ablation}
\vspace{-2mm}
\end{table}}

Given the lack of generative methods that can scale up to this data,
we compare with a monocular depth method from \cite{sax2020visualpriors}, also trained on Taskonomy. We first adjust the absolute scale and shift of its output by comparing with ground truth depth (as in the loss function of MiDaS \cite{lasinger2019midas}) and proceed by unprojecting with the same procedure as when creating the dataset. For \ours{}, we directly predict the point clouds in absolute units, using the $(u,v,l)$ reparameterization introduced in Sec.~\ref{sec:method-conditioning}. We use 2048 points for both training and evaluation. We compare the two approaches qualitatively in Fig.~\ref{fig:omnidata}, and quantitatively in Table~\ref{tab:taskonomy},
using the same metric as in Sec.~\ref{sec:experiments-shapenet-conditional} and Table~\ref{tab:image-conditional}. This experiment confirms our method extends beyond object-centric views to real scenes, and greatly outperforms similarly-sized baselines benefitting from years of research on monocular depth.
As in Sec~\ref{sec:experiments-shapenet-conditional}, we also report results with ICP. For the baseline we disabled scale estimation, as it degrades the results.

\subsection{Ablation studies and further experiments}
\label{sec:experiments-ablation}

\para{Ablation study: Global vs projective conditioning}
We evaluate our approach to conditioning the denoising process through projective geometry with the more standard approach relying on a global embedding. Instead of bilinear lookup for each point, we mean-pool the CNN features and inject them globally as in \cite{zeng2022lion}, alongside $t$, through the bias and scale of normalization layers. We compare both approaches on the OccNet benchmark of Sec.~\ref{sec:experiments-shapenet-conditional}.
As we do not aim to compare against those baselines,
we use 1024 points rather than 100k, for simplicity. Results are shown in Table~\ref{tab:ablation}.
Our projective conditioning is 7-8\% more accurate.

\para{Ablation study: Numerical solvers} We consider the OccNet \cite{onet} benchmark and ablate the use of the probability flow solver proposed in \cite{song2021scorebased} versus the stochastic solver used in \cite{karras2022elucidating}. We find the latter superior and turn to the effect of the number of solver steps. We observe that more solver steps bring larger perceptual improvements which are not always captured by the chamfer distance, up to about 128 steps, which we use in all other experiments in the paper. \et{We report the numbers in Table~\ref{tab:ablation}: relative values take $N_{denoise}=128$ as reference.} It should be noted that technically, both solvers use two network evaluations per step.

\et{\para{Ablation study: training with fewer samples} We train conditional models with only 50\% of the data and report a surprisingly small drop in performance: $\mytilde$2\% (Table~\ref{tab:ablation}). This holds for both projective and global conditioning.}

\para{Probabilities as a metric}
Our approach allows us to compute exact probabilities over the validation set, following the method in \cite{song2021scorebased}. The maxima in probability corresponds to the optimal state in terms of \textit{novel} generative performance. On ShapeNet, which is quite small, we notice that our models overfit in terms of this metric while maintaining low 1-NN accuracy: we report their evolution in the unconditional setting in Fig.~\ref{fig:logp}. This corroborates our observation that the ShapeNet benchmark is relatively saturated: by some metrics, state-of-the-art methods may even outperform an oracle which simply returns the training set. We argue that tractable likelihoods may prove very useful in datasets and tasks with no other easy means of validation.
\begin{figure}
\centering
\includegraphics[width=\linewidth]{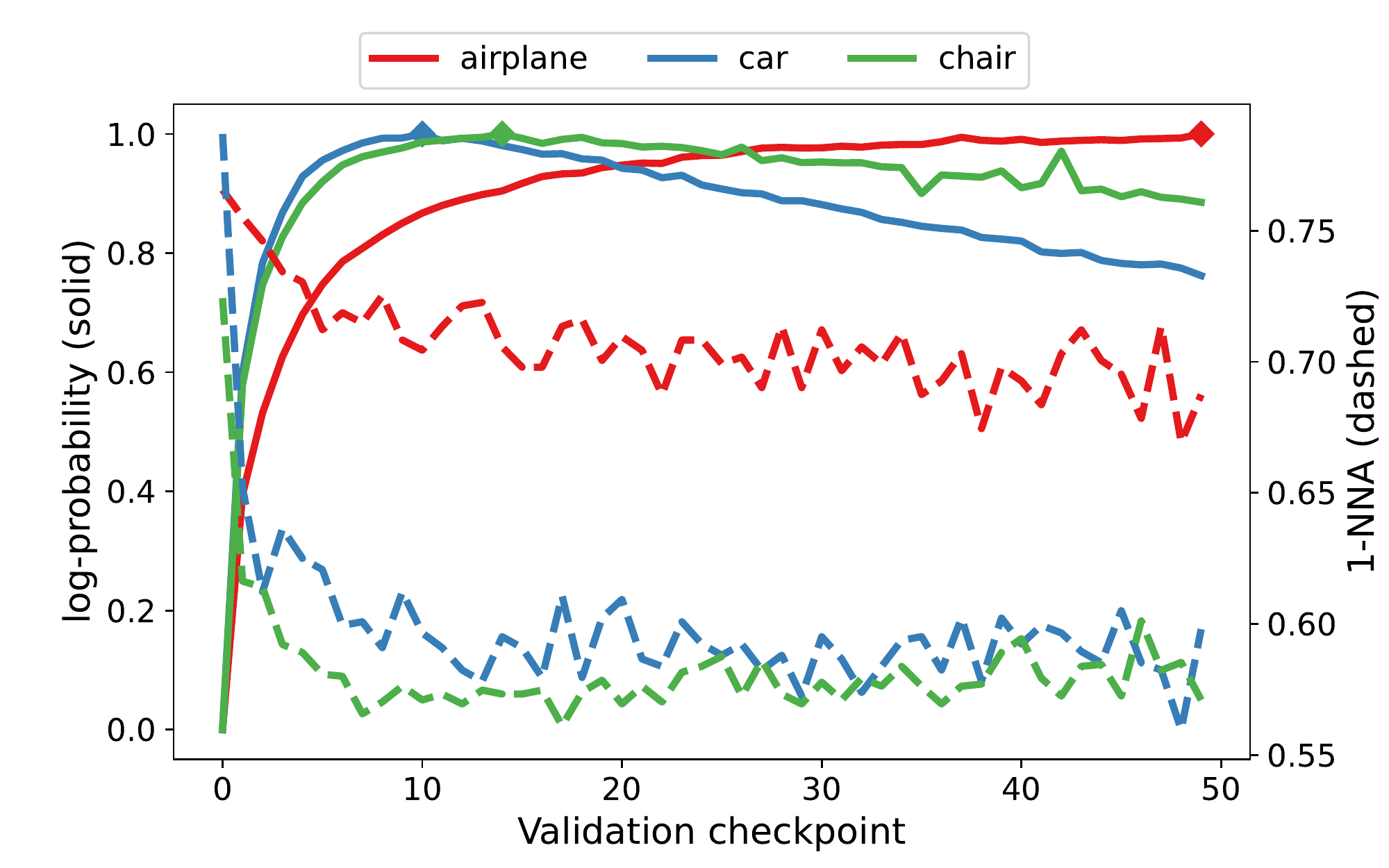}
\caption{{\bf Log-probabilities while training.} Dashed line: CD 1-NNA (Sec.~\ref{sec:experiments-shapenet-unconditional}) on the validation set. Solid: validation set log-probability (normalized); diamonds mark maxima. Notice how the decreasing likelihood of sampling validation examples does not increase (\ie, degrade) 1-NNA.}
\vspace{-3mm}
\label{fig:logp}
\end{figure}

\para{Point cloud upsampling by inpainting}
Another application of our method is point cloud upsampling. To upsample from $n$ to $m$ points, following the blueprint of \cite{lugmayr2022repaint}, we diffuse input $\set{p_{0..n}}$ to $\sigma_\mathrm{max}$ and concatenate with new points sampled from the prior, $\set{p_{n..m}}$. We then reverse-diffusion, computing the scores on all points $\set{p_{0..m}}$, but use them only to update $\set{p_{n..m}}$, while $\set{p_{0..n}}$ is reversed deterministically to the input. Compared to models following Eq.~\ref{eq:latent-shape}, this approach treats the input $\set{p_{0..n}}$ as the latent code $s$. We use 4 resampling sub-steps (see \cite{lugmayr2022repaint}) per solver step, and when upsampling by large factors, in order to stay in the range of $m$ the network is trained for, we concatenate multiple conditionally-independent completions of $\set{p_{0..n}}$. We find this procedure to result in coherent, high-resolution point clouds: see Fig.~\ref{fig:inpainting-examples} for examples.

\renewcommand{\width}{1.6cm}
{\renewcommand{\arraystretch}{0.3}
\setlength{\tabcolsep}{0.5pt}
\begin{figure}
\centering
\definecolor{LightGray}{rgb}{0.85, 0.85, 0.85}
\newcolumntype{?}[1]{!{\color{LightGray}\vrule width #1}}
\begin{tabu}{@{}ccc?{1pt}cc@{}}
\includegraphics[width=\width]{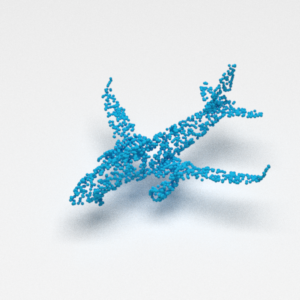} &
\includegraphics[width=\width]{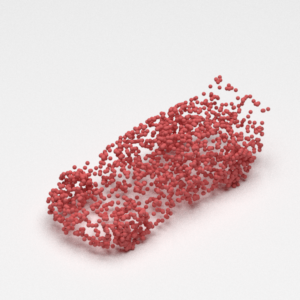} &
\includegraphics[width=\width]{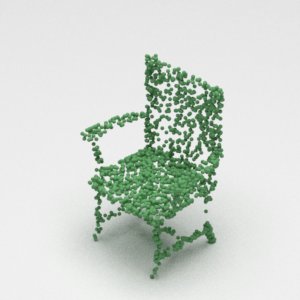}\hspace{1mm} &
\hspace{1mm}
\includegraphics[width=\width]{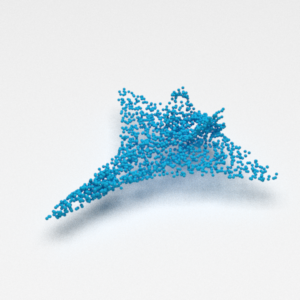} &
\includegraphics[width=\width]{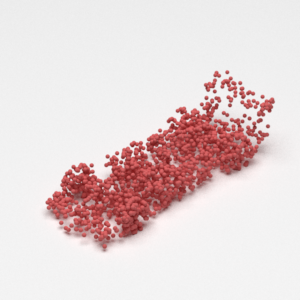} \\
\includegraphics[width=\width]{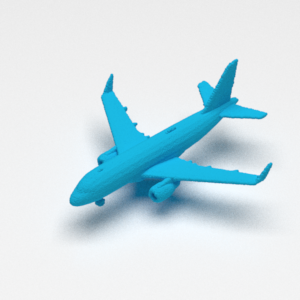} &
\includegraphics[width=\width]{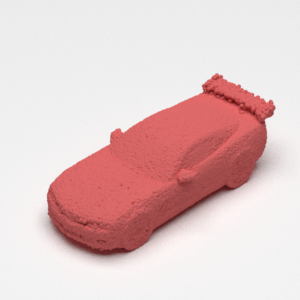} &
\includegraphics[width=\width]{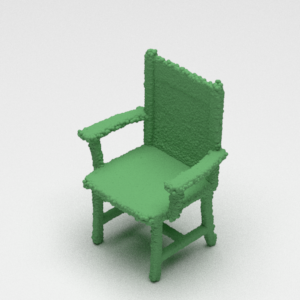}\hspace{1mm} &
\hspace{1mm}\includegraphics[width=\width]{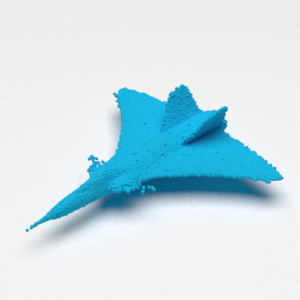} &
\includegraphics[width=\width]{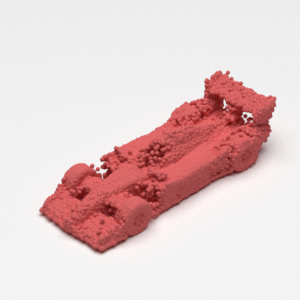} \\
\end{tabu}
\vspace{-2mm}
\caption{{\bf Point cloud upsampling examples.} Top: original point clouds with 2048 points. Bottom: we upsample them by 50x to 102k points. Left: Our inpainting technique yields high-quality point clouds. Right: Artifacts may appear occasionally, especially on complex, irregular structures. Note: figure is rendered with a smaller point size.}
\label{fig:inpainting-examples}
\vspace{-1mm}
\end{figure}}

{\renewcommand{\arraystretch}{0.9}
\setlength{\tabcolsep}{6pt}
\begin{table}
\small
\centering
\begin{tabular}{lcc}
\toprule
& Num. params & Inference speed \\
\midrule
LION (unconditional) & 110M & 2.51 s/example \\
PVD (unconditional) & 27.7M & 2.95 s/example \\
\ours{} (unconditional) & 13.7M & 0.25 s/example \\
\midrule
\ours{} (conditional) & 47.7M & 0.27 s/example \\
\bottomrule
\end{tabular}
\vspace{-2mm}
\caption{{\bf Size and speed comparison.} Measured on an NVIDIA A100 GPU with 40Gb, with 2048 points and batch of 64. \ours{} uses the SDE sampler with 128 steps.}
\label{tab:runtime}
\vspace{-2mm}
\end{table}}
\section{Conclusions}

\et{We propose a novel approach to condition denoising diffusion models in a geometrically-principled manner by projecting generated point clouds to an image and augmenting point locations with sparse features from a convolutional backbone.
Our framework relies on a simple permutation-equivariant transformer, trained with a continuous-time diffusion scheme.
It yields state-of-the-art results in single-view synthesis, while performing on par
in the unconditional setting.
It can also deliver exact probabilities, and upsample by inpainting. We believe this is a first step towards controllable diffusion point cloud models on real data.
Future work will explore occlusion on large-scale datasets,
multi-view inference,
and completion via inpainting.}

\paragraph{Acknowledgements} This research was partially funded by Google’s {\it Visual Positioning System.}

\bigskip
\clearpage
\begin{center}
  \LARGE Supplementary Material
\end{center}
\setcounter{section}{0}
\renewcommand{\thesection}{\Alph{section}}

\section{\mt{Experimental} details}
\mt{For simplicity we measure training in terms of gradient updates rather than dataset epochs.}

\mt{\paragraph{Unconditional ShapeNet~\cite{yang2019pointflow}.} We use the download links provided by the PointFlow paper~\cite{yang2019pointflow} directly. \et{Their ShapeNet models contain 100k points: we subsample them to 2048 points while training, as explained in the paper.} We train our models for 500k steps, with batch size of 64, and evaluate the model every 10k steps. For this evaluation we run a simplified variant of the benchmark described in section 4.1, which uses only the chamfer distance to pick the best checkpoint \et{over the validation set} -- a variant of early stopping.}

\mt{\paragraph{ShapeNet-vol~\cite{onet}.} We use the download links provided by the authors of OccNet~\cite{onet}, \et{which contain prerendered images in addition to point clouds}. \et{Their point clouds are already subsampled to 2048 points.} We train our models for 500k steps, with batch size of 48 and evaluate the model every 10k steps with a simplified version of the benchmark described in 4.2 not using ICP and with a random subset of the validation set. Since we observed no overfitting, we do not do early stopping.}

\mt{\paragraph{Taskonomy~\cite{zamir2018taskonomy}.} The dataset comes in 4 sizes: \emph{tiny}, \emph{medium}, \emph{full} and \emph{full-plus}. We noticed that \emph{full-plus} contains certain scenes with incomplete/faulty scans. We train our model for 1M steps on the \emph{full} variant and evaluate it on a randomly sampled 1024-element subset of the `validation' split every 10k steps - the subsampling is to save computation. Note that the monocular depth baseline of \cite{sax2020visualpriors} was also trained (by the authors) on the Taskonomy dataset.

We use the same evaluation procedure as for ShapeNet-vol, this time with 8192 examples, and observe no overfitting, so we report numbers for the final checkpoint. This means that the results over `validation' and over `test' should be equivalent, up to uniformity in the dataset split, \et{and report both in the paper}.

\et{However, we notice that performance is lower on the `test' split, particularly in terms of the mean, and not the median.} We trace this discrepancy to two out-of-distribution scenes in the `test' subset: `vacherie', a residential apartment with very high ceilings which GECCO mis-scales by a factor of $2 \times$, and `german', a gym filled with large windows and mirrors, resulting in ambiguous depth (see Fig.~\ref{fig:taskonomy-outliers}). Note that the monocular baseline predicts \emph{relative} depth, which we then rescale using the ground truth, which makes it largely immune to such outliers.
\et{In fact, the large difference in performance between GECCO without ICP and with ICP is mostly explained by mis-estimating the scale of the scene, which the baseline doesn't suffer from, due to its use of ground truth scale. This is also likely the reason why the baseline performs worse if ICP estimates both scale and rotation -- as explained in the paper, we disable scale estimation for it.}
}

\section{Unconditional evaluation metrics}
\newcommand{\Sr}[0]{\mathbf{S_r}}
\newcommand{\Sg}[0]{\mathbf{S_g}}
\newcommand{\Sa}[0]{\mathbf{S_a}}
\newcommand{\Sb}[0]{\mathbf{S_b}}

Here we define the metrics of coverage ($\mathrm{COV}$), minimum matching distance ($\mathrm{MMD}$) and 1-nearest neighbour accuracy ($\mathrm{1-NNA}$). They are equivalent to those used in \cite{yang2019pointflow}, but using a different notation. We assume  $n$ given reference point clouds from the test set $\Sr = \set{p_1, ..., p_n}$, $n$ samples generated by the model $\Sg = \set{q_1, ..., q_n}$, and a distance function $D(\cdot, \cdot)$ which can be either CD or EMD (defined in main text).

\paragraph{Coverage} considers the distances between $p \in \Sr$ and $q \in \Sg$ and measures the fraction of $p$ which are the nearest neighbour of some $q$:
\begin{multline}
    \mathbf{COV}(\Sr, \Sg) = \frac{1}{n} \\
        \left|\left\{ q \in \Sr : \left( \underset{p \in \Sg}{\exists} q = \underset{\hat{q} \in \Sr}{\argmin} D(\hat{q}, p) \right) \right\}\right|.
\end{multline}

\paragraph{Mean matching distance} measures the average distance between a sample and its nearest reference point
\begin{equation}
    \mathbf{MMD}(\Sr, \Sg) = \frac{1}{n} \underset{p \in \Sr}{\sum} \underset{q \in \Sg}{\min} D(p, q).
\end{equation}

\paragraph{1-nearest neighbour accuracy} measures the accuracy of a classifier distinguishing between elements of $\Sr$ and $\Sg$ by returning the class of the nearest example in $\Sr \cup \Sg - \set{u}$, where $u$ is the element under consideration itself. We define this formally by introducing the asymmetric $H(\Sa, \Sb)$ which counts the $a \in \Sa$ correctly classified w.r.t. the union of $\Sa$ and $\Sb$ and then define $\mathbf{1-NNA}$ as the average of $H$ applied in both directions. We introduce $\Sa, \Sb, u$ and $v$ to emphasize that in the final equation $H$ is applied in both directions.
\begin{equation}
    H(\Sa, \Sb) = \left|\left\{ u \in \Sa : \left[ \underset{v \in (\Sa \cup \Sb - \set{u})}{\argmin} D(u, v) \in \Sa \right] \right\}\right|
\end{equation}
\begin{equation}
    \mathbf{1-NNA}(\Sr, \Sg) = \frac{1}{2n} \left(H(\Sr, \Sg) + H(\Sg, \Sr)\right)
\end{equation}

\mt{We note that \et{EMD is usually approximated due to its computational cost, and that} CUDA-based implementations may vary drastically; we use \href{https://github.com/daerduoCarey/PyTorchEMD}{PyTorchEMD}.}

\section{Additional visualizations}
We invite the reader to view the video included in this supplementary material for conditional and unconditional samples from ShapeNet, as well as for image-conditional samples from Taskonomy. On the following pages of this document, in Fig.~\ref{fig:supp:unconditional} we show additional unconditional samples from models trained on the ShapeNet, compared with baselines~\cite{luo2021diffusion,cai2020learning}. In Fig.~\ref{fig:supp:taskonomy-upsampling} we show additional samples from the image-conditional model trained on the Taskonomy dataset, along with the results of their upsampling. Note that we upsample the point clouds with the technique introduced in sec. 4.4 of the main paper, and how showcase it in the image-conditional case. Due to performance issues with the point cloud renderer we render the results upsampled only by 5x, but GECCO itself works equally well with 100k points, as shown in the main paper.

\mt{\section{Efficient upsampling}
During upsampling by inpainting, for optimal performance, it is preferable to keep the network's input distribution as close as possible to that at training time.
\et{This means that ideally, we would want $m = n + 1$, where $n$ and $m$ are the cardinality of the original and upsampled point clouds, respectively.}
Note that a cloud can be upsampled by a factor of $f$, to $(1 + f)n$ points, by repeating that procedure $f \times n$ times and concatenating the results. This however is as expensive as generating a cloud of $n$ points $f \times n$ times from scratch.

Fortunately, the Set Transformer~\cite{lee2019set} architecture enables a workaround which brings the cost down to that of sampling a cloud of $f \times n$ points once. Since the points do not interact directly with each other, but rather via the inducers (see the main paper and \cite{lee2019set} for definition), if we make the assumption that in the limit of large $n$ the influence of any single point on the inducers is negligible, we can reverse-diffuse many new (inpainted) points \emph{in parallel}, with shared inducer state. This leads to a simple algorithm starting from the conditioning set $\set{p}_{1..n}^{\sigma = 0}$ and the pure noise inputs $\set{p}_{n..m}^{\sigma = \sigmamax}$. At each step $t$ of reverse diffusion:
\begin{enumerate}
    \item The conditioning input is diffused to $\sigma_t$ to obtain $\set{p}_{1..n}^{\sigma_t}$.
    \item The score network is evaluated on $\set{p}_{1..n}^{\sigma_t}$. The score estimate is discarded and instead we cache the activations of the inducers across the network's layers.
    \item The score network is ran again, this time on $\set{p}_{n..m}^{\sigma_t}$, but with the inducer activations provided by the cache from point 2. We obtain the score for the inpainted points $n..m$.
    \item We update $\set{p}_{n..m}$ as usual, using the score from point 3.
\end{enumerate}

The results provided in the main paper and here are obtained by the naive procedure, upsampling multiple times with a context of $\frac{1}{2}n = 1024$ but we also release code implementing the improved scheme described here.}

\section{Dataset licenses}
\paragraph{ShapeNet.} The original ShapeNet~\cite{chang2015shapenet} dataset of 3D meshes is \href{https://shapenet.org/terms}{licensed} for non-commercial use, with clauses allowing for redistribution of derived assets. Our use of derived datasets released with \cite{yang2019pointflow} and \cite{onet} fulfills these provisions.\\

\paragraph{Taskonomy.} Our dataset of real world imagery is derived from the data released with Taskonomy~\cite{zamir2018taskonomy}, which allows non-commercial use in its \href{https://github.com/StanfordVL/taskonomy/blob/9f814867b5fe4165860862211e8e99b0f200144d/data/LICENSE}{license}. Bound by the license, we are not able to release our derivatives until authorized by the authors of \cite{zamir2018taskonomy}. Depending on their permission, we intend to release either our dataset or the code to re-generate it from the original files Taskonomy files.

\renewcommand{\width}{1mm}
\renewcommand{\height}{1.6cm}

{

\begin{figure*}
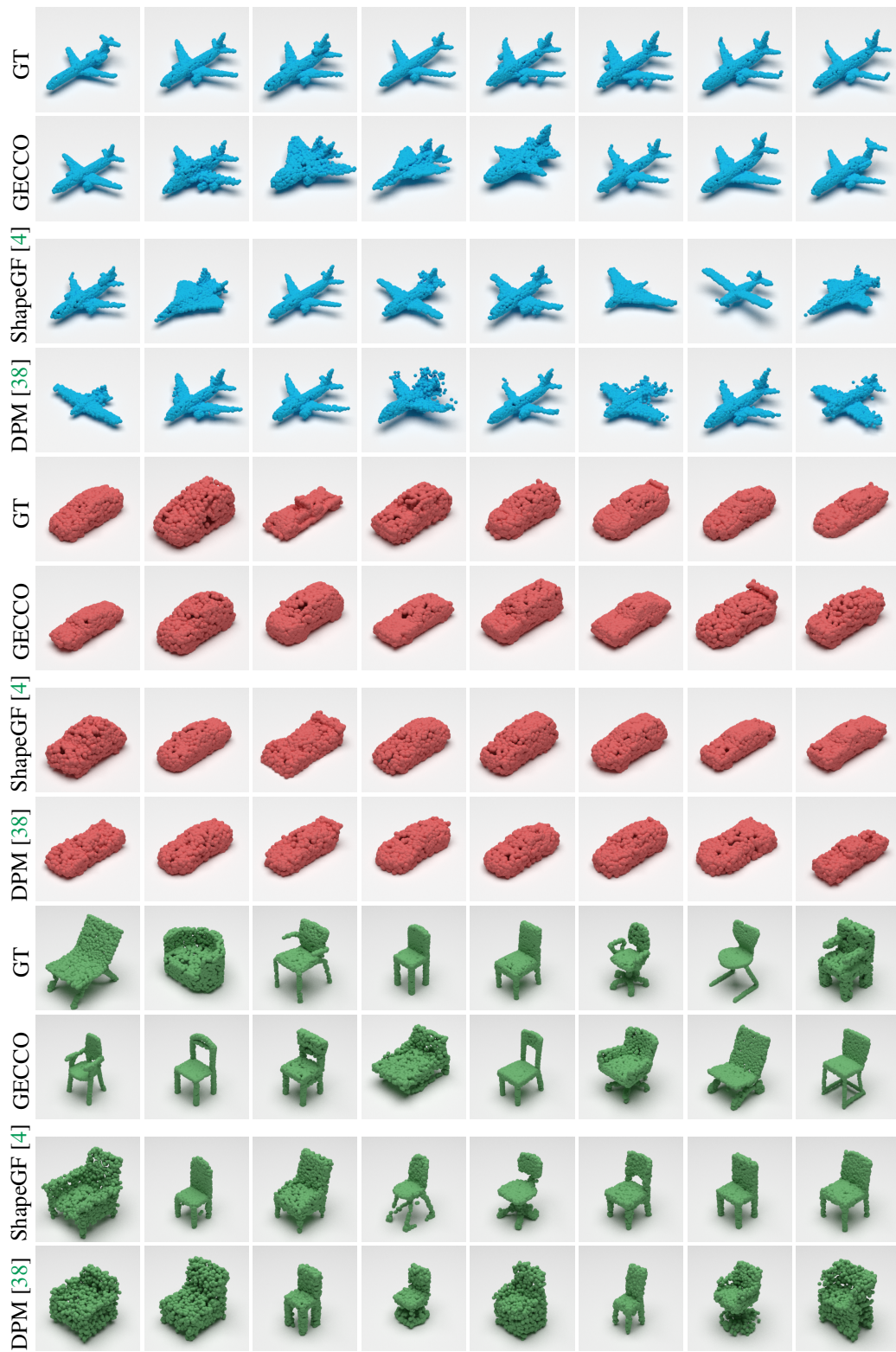

\centering

\newcommand{\imageRow}[3]{
\rotatebox{90}{#3} &
\includegraphics[height=\height]{figures/unconditional/output/#1-#2-00} &
\includegraphics[height=\height]{figures/unconditional/output/#1-#2-01} &
\includegraphics[height=\height]{figures/unconditional/output/#1-#2-02} &
\includegraphics[height=\height]{figures/unconditional/output/#1-#2-03} &
\includegraphics[height=\height]{figures/unconditional/output/#1-#2-04} &
\includegraphics[height=\height]{figures/unconditional/output/#1-#2-05} &
\includegraphics[height=\height]{figures/unconditional/output/#1-#2-06} &
\includegraphics[height=\height]{figures/unconditional/output/#1-#2-07} \\
}

\newcommand{\categoryComparison}[1]{
\imageRow{ground_truth}{#1}{\quad\, GT}
\imageRow{pox}{#1}{\, GECCO}
\imageRow{shape_gf}{#1}{ShapeGF~\cite{cai2020learning}}
\imageRow{dpm}{#1}{DPM~\cite{luo2021diffusion}}
}

\setlength{\tabcolsep}{1pt}
\begin{tabular}{@{}ccccccccc@{}}

\categoryComparison{airplane}
\categoryComparison{car}
\categoryComparison{chair}

\end{tabular}

\caption{{\bf Unconditional point cloud synthesis on ShapeNet.} All examples contain 2048 points.}
\label{fig:supp:unconditional}
\end{figure*}}
\renewcommand{\width}{1mm}
\renewcommand{\height}{3cm}

{

\begin{figure*}
\centering

\newcommand{\imageRow}[2]{
    \rotatebox{90}{#2} &
    \includegraphics[height=\height]{sections/supp/taskonomy-upsampling/frames/#1_0.png} &
    \includegraphics[height=\height]{sections/supp/taskonomy-upsampling/frames/#1_1.png} &
    \includegraphics[height=\height]{sections/supp/taskonomy-upsampling/frames/#1_2.png} &
    \includegraphics[height=\height]{sections/supp/taskonomy-upsampling/frames/#1_3.png} &
    \includegraphics[height=\height]{sections/supp/taskonomy-upsampling/frames/#1_4.png} \\
}

\setlength{\tabcolsep}{1pt}
\begin{tabular}{@{}cccccc@{}}

\imageRow{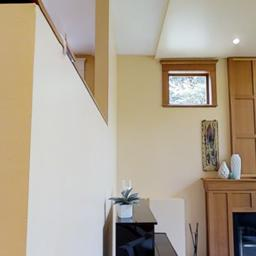}{\qquad\; Image}
\imageRow{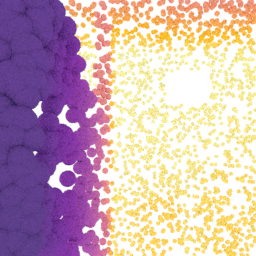}{\qquad GT (2048)}
\imageRow{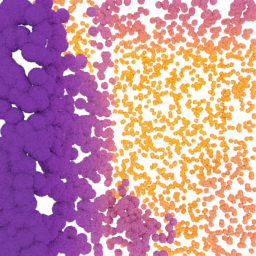}{\quad GECCO (2048)}
\imageRow{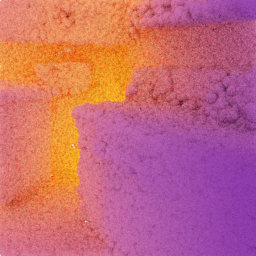}{\; GECCO (10240)}
\imageRow{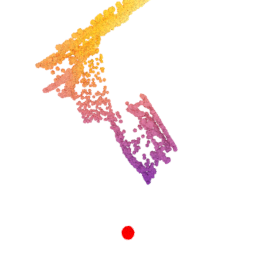}{\qquad GT (2048)}
\imageRow{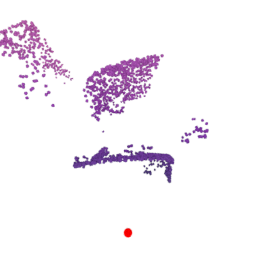}{\qquad GECCO (2048)}
\imageRow{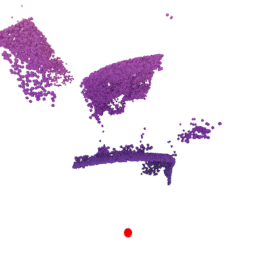}{\; GECCO (10240)}

\end{tabular}

\caption{{\bf Image-conditional point cloud synthesis and upsampling on the test set of Taskonomy.} We render the point clouds with smaller point radii, to highlight the difference made by upsampling. Note how in column 4 GECCO is tricked by a mirror, \et{on the left side}.}
\label{fig:supp:taskonomy-upsampling}
\end{figure*}}
\renewcommand{\height}{1.9cm}
\DeclareUrlCommand\UScore{\urlstyle{rm}}
\newcommand{\vpscene}[1] {\taskonomyscene{sections/supp/taskonomy-outliers/vp-top/#1}}
\newcommand{\geccoscene}[1] {\taskonomyscene{sections/supp/taskonomy-outliers/gecco-top/#1}}

{\renewcommand{\arraystretch}{0.6}
\begin{figure*}
\centering
\setlength{\tabcolsep}{1pt}
\begin{tabular}{@{}cccccccc@{}}
\vpscene{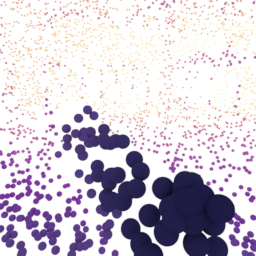}\\
\vpscene{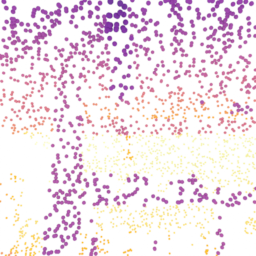}\\
\vpscene{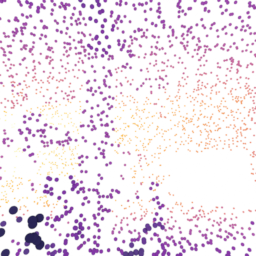}\\
\vpscene{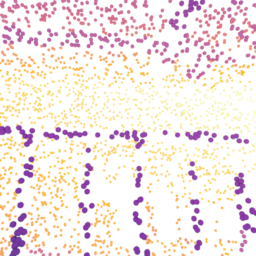}\\
\vpscene{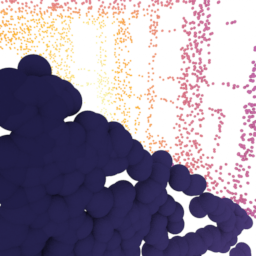}\\
\\
\hline\\
\geccoscene{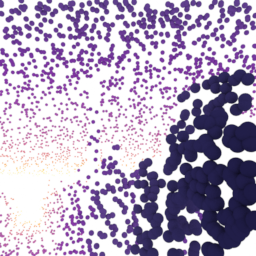}\\
\geccoscene{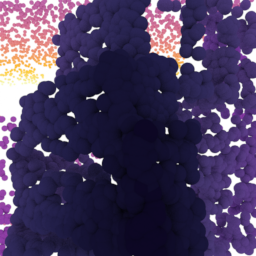}\\
\geccoscene{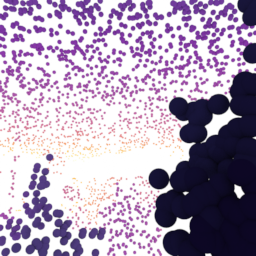}\\
\geccoscene{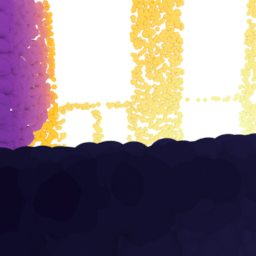}\\
\geccoscene{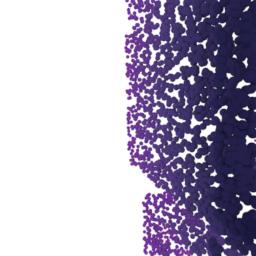}\\
{\footnotesize (a) Source image} & {\footnotesize (b) GT depth} & {\footnotesize (c) GT POV} & {\footnotesize (d) \ours{} POV} & {\footnotesize (e) \cite{sax2020visualpriors} POV} & {\footnotesize (f) GT BEV} & {\footnotesize (g) \ours{} BEV} & {\footnotesize (h) \cite{sax2020visualpriors} BEV} \\
\end{tabular}
\caption{{\bf Outliers in Taskonomy.} We showcase the top 5 test images where the difference in chamfer distance is most in favor of the monocular baseline (top) and \ours~(bottom). GECCO suffers from mirrors (top 4 rows, `german') and unusual room dimensions (last row, `vacherie'), while the baseline cannot resolve areas of undefined depth, such as in the two bottom rows. }
\label{fig:taskonomy-outliers}
\end{figure*}}

\section{Emoji license}
In the \mt{video material} we use an emoji (\includegraphics[height=4mm,trim=0mm 10mm 0mm -10mm]{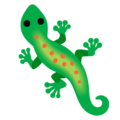}) from the \href{https://github.com/googlefonts/noto-emoji/tree/934a5706f1f3dd2605c4d9b5d9162fd7f89d8702}{Noto emoji bank}, licensed under Apache 2.0 license.

\clearpage

{\small
\bibliographystyle{ieee_fullname}
\bibliography{egbib}
}

\end{document}